\documentclass[sigconf]{acmart}
\usepackage{multirow}
\usepackage{mathrsfs}
\usepackage{makecell}
\usepackage{hyperref}
\AtBeginDocument{%
  \providecommand\BibTeX{{%
    \normalfont B\kern-0.5em{\scshape i\kern-0.25em b}\kern-0.8em\TeX}}}

\setcopyright{acmcopyright}
\copyrightyear{2018}
\acmYear{2018}
\acmDOI{10.1145/1122445.1122456}

\acmConference[Woodstock '18]{Woodstock '18: ACM Symposium on Neural
  Gaze Detection}{June 03--05, 2018}{Woodstock, NY}
\acmBooktitle{Woodstock '18: ACM Symposium on Neural Gaze Detection,
  June 03--05, 2018, Woodstock, NY}
\acmPrice{15.00}
\acmISBN{978-1-4503-XXXX-X/18/06}

\begin{document}
% \fancyhead{}

\title{Temporal Knowledge Graph Reasoning Based on Evolutional Representation Learning}
%Event Knowledge Graph based Evolving Embedding Learning

\author{Zixuan Li\textsuperscript{1,2}, Xiaolong Jin\textsuperscript{1,2}, Wei Li\textsuperscript{3}, Saiping Guan\textsuperscript{1,2}, Jiafeng Guo\textsuperscript{1,2}, Huawei Shen\textsuperscript{1,2}, Yuanzhuo Wang\textsuperscript{1,2} and Xueqi Cheng\textsuperscript{1,2}}
\affiliation{
  \institution{
    \textsuperscript{1}School of Computer Science and Technology, University of Chinese Academy of Sciences, Beijing, China}
  \institution{
    \textsuperscript{2}CAS Key Lab of Network Data Science and Technology, ICT, CAS, Beijing, China\\
    }
  \institution{
    \textsuperscript{3}Baidu Inc., Beijing, China}
    \country{} 
}
\email{{lizixuan,jinxiaolong,guansaiping,guojiafeng,shenhuawei,wangyuanzhuo,cxq}@ict.ac.cn; liwei85@baidu.com}


\renewcommand{\shortauthors}{Li. and Jin., et al.}

\begin{abstract}
Knowledge Graph (KG) reasoning that predicts missing facts for incomplete KGs
has been widely explored. However, reasoning over Temporal KG (TKG) that
predicts facts in the future is still far from resolved. The key to predict
future facts is to thoroughly understand the historical facts. A TKG is actually
a sequence of KGs corresponding to different timestamps, where all concurrent
facts in each KG exhibit structural dependencies and temporally adjacent facts
carry informative sequential patterns. To capture these properties effectively
and efficiently, we propose a novel Recurrent Evolution network based on Graph
Convolution Network (GCN), called RE-GCN, which learns the evolutional
representations of entities and relations at each timestamp by modeling the KG
sequence recurrently. Specifically, for the evolution unit, a relation-aware GCN
is leveraged to capture the structural dependencies within the KG at each
timestamp. In order to capture the sequential patterns of all facts in parallel,
the historical KG sequence is modeled auto-regressively by the gate recurrent
components. Moreover, the static properties of entities such as entity types,
are also incorporated via a static graph constraint component to obtain better
entity representations. Fact prediction at future timestamps can then be
realized based on the evolutional entity and relation representations. Extensive
experiments demonstrate that the RE-GCN model obtains substantial performance
and efficiency improvement for the temporal reasoning tasks on six benchmark
datasets. Especially, it achieves up to 11.46\% improvement in MRR for entity
prediction with up to 82 times speedup comparing to the state-of-the-art
baseline.

\end{abstract}


\maketitle
% \vspace{-1.5mm}
\section{Introduction}
  
Knowledge Graphs (KGs) have facilitated many real-world
applications~\cite{zou2020survey}. However, they are usually incomplete, which
restricts the performance and range of KG-based applications. To alleviate this
problem, reasoning over KG~\cite{bordes2013translating,trouillon2016complex}
that attempts to predict missing facts, is a critical task in natural language
processing. Traditionally, a KG is considered to be static multi-relational
data. However, recent availability of large amount of event-based interaction
data~\cite{boschee2015icews} that exhibits complex temporal dynamics has created
the need for approaches that can characterize and reason over Temporal Knowledge
Graph (TKG)~\cite{boschee2015icews,gottschalk2018eventkg,gottschalk2019eventkg}.
A fact in a TKG can be represented in the form of (subject entity, relation,
object entity, timestamp). Actually, a TKG can be denoted as a sequence of KGs
with timestamps, each of which contains the facts that co-occur at the same
timestamp. The left part of Figure~\ref{fig:ekg_and_tasks} illustrates an
example of TKG from the ICEWS18~\cite{jin2020Renet} dataset. Despite the
ubiquitousness of TKGs, the methods for reasoning over such kind of data are
relatively unexplored both in effectiveness and efficiency.

Reasoning over a TKG from timestamps $t_{0}$ to $t_T$ primarily has two
settings, interpolation and extrapolation~\cite{jin2020Renet}. The
former~\cite{dasgupta2018hyte,garcia2018learning, leblay2018deriving} attempts
to infer missing facts from $t_{0}$ to $t_T$~\cite{jin2020Renet}. The
latter~\cite{jin2020Renet,jin2019recurrent,trivedi2017know,trivedi2018dyrep},
which aims to predict future facts (events) for time $t > t_{T}$, is much more challenging. 
For TKG, predicting new facts at
future timestamps based on the observed historical KGs is helpful for understanding
the hidden factors of events and responding to emerging
events~\cite{muthiah2015planned, phillips2017using, korkmaz2015combining}. Thus
reasoning under the extrapolation setting is very vital and can be helpful for
many practical applications, such as disaster relief~\cite{signorini2011use} and
financial analysis~\cite{bollen2011twitter}. In this paper, the temporal
reasoning tasks (i.e., reasoning under the extrapolation setting over TKGs)
contains two subtasks as shown in the right part of Figure~\ref{fig:ekg_and_tasks}:

\begin{figure}[htbp]
  \centering
  \includegraphics[width=3in]{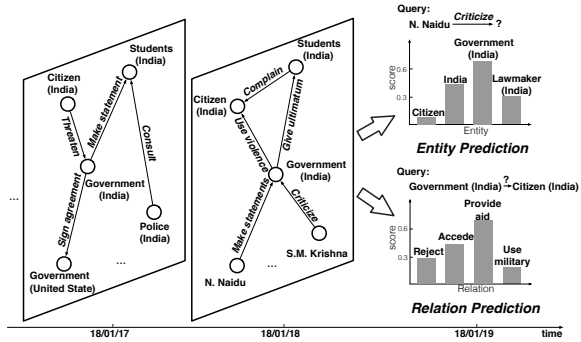}
  \caption{An illustration of temporal reasoning over a TKG. We present two
  subgraphs from the KGs at timestamps 18/01/17 and 18/01/18, respectively. %The temporal reasoning task has two settings, i.e., entity prediction and relation prediction.
  }
  \label{fig:ekg_and_tasks}
  \vspace{-5mm}
  \end{figure}

\begin{itemize}
\item {\bf Entity Prediction}: Predict which entity will have a given relation
together with a given entity at a certain future timestamp (e.g., Who will
N.Naidu criticize at 18/01/19?);
\item {\bf Relation Prediction}: Predict the relation that will occur between
two given entities at a certain future timestamp (e.g., What will happen between
Government (India) and citizen (India) at 18/01/19?).
\end{itemize}

To accurately predict future facts, it requires the model to dive deeply into
historical facts. At each timestamp, entities influence each other via
concurrent facts, which form a KG and exhibit complex \textbf{structural
dependencies}. As an example shown in Figure~\ref{fig:ekg_and_tasks}, the
concurrent facts at \emph{18/01/18} demonstrates that \emph{Government (India)}
is under pressure from many people, which may influence the behaviors of
\emph{Government (India)} at 18/01/19. Besides, the behaviors of each entity
embodied in temporally adjacent facts may carry informative \textbf{sequential
patterns}. As shown in Figure~\ref{fig:ekg_and_tasks}, the historical behaviors
of \emph{N. Naidu} reflect his preferences and affect his future behaviors to a
certain degree. The combination of these two kinds of historical information
drives the behavioral trends and preferences of entities and relations.

Some earlier attempts including Know-evolve~\cite{trivedi2017know} and its
extension DyRep~\cite{trivedi2018dyrep}, learn evolutional entity
representations by modeling the occurrence of all facts in the history as a
temporal point process. However, they can not model concurrent facts at the same
timestamps. Some recent attempts extract some related historical information for
each individual query in a heuristic manner. Specifically,
RE-NET~\cite{jin2019recurrent, jin2020Renet} extracts those directly-engaged
historical facts for the given entity in each query of entity prediction and
then encodes them sequentially. CyGNet~\cite{zhu2020learning} models the
historical facts with the same entity and relation to each query of entity
prediction, and thus mainly focuses on predicting facts with repetitive
patterns. As a TKG is actually a KG sequence, the existing methods have three
main restrictions: (1) mainly focusing on the entity and relation of a given
query and neglecting the structural dependencies among all the facts in the KG
at each timestamp; (2) low efficiency by encoding the history for each query
individually; (3) ignoring the function of some static properties of entities such
as entity types. Besides, the existing methods only focus on entity prediction,
while relation prediction cannot be solved simultaneously by the same model.

In this work, we treat TKG as a KG sequence and model the whole KG sequence
uniformly to encode all historical facts into entity and relation
representations to facilitate both entity and relation prediction tasks. Thus,
we propose a novel GCN-based Recurrent Evolution network, namely RE-GCN, which
learns the evolutional representations of entities and relations at each
timestamp by modeling the KG sequence recurrently. Specifically, for each
evolution unit, a relation-aware GCN is leveraged to capture the structural
dependencies within the KG at each timestamp. In this way, the interactions
among all the facts in a KG can be effectively modeled. The historical KG
sequence is modeled auto-regressively by the gate recurrent components to
capture the sequential patterns across all temporally adjacent facts
efficiently. All the historical information of entities and relations in the TKG
are encoded in parallel. Moreover, the static properties of entities, such as
entity types, are also incorporated via a static-graph constraint component to
obtain better entity representations. Then, the tasks of entity prediction and
relation prediction at future timestamps can be realized based on the
evolutional representations.

In general, this paper makes the following contributions:
 \begin{itemize}

\item We propose an evolutional representation learning model RE-GCN for
temporal reasoning over TKGs, which considers the structural dependencies among
concurrent facts in a KG, the sequential patterns across temporally adjacent
facts, and the static properties of entities. To the best of our knowledge, this is
the first study that integrates all of them into the evolutional representations
for temporal reasoning.
\item By characterizing TKG from the view of a KG sequence, RE-GCN efficiently
models all the historical information in the TKG into evolutional
representations, which are applicable for both entity
and relation prediction simultaneously. Therefore, it enables up
to 82 times speedup compared to the state-of-the-art baseline.
\item Extensive experiments demonstrate that, by modeling the history more
comprehensively, RE-GCN achieves consistently and significantly better
performance (up to 11.46\% improvement in MRR) over both entity and relation
prediction tasks on six commonly used benchmarks.
\end{itemize}

\begin{figure*}[htbp]
\centering
\includegraphics[width=6in]{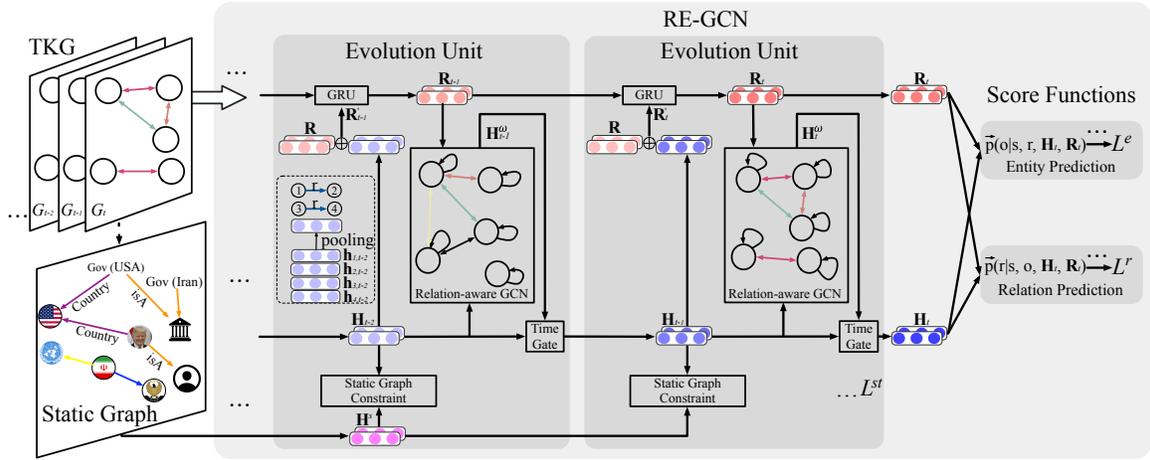}
\caption{An illustrative diagram of the proposed RE-GCN model for temporal
reasoning at timestamp $t+1$.}
% The input of RE-GCN is the historical KG sequence
% in the TKG and the derived static graph. At each timestamp, the evolution unit
% updates the entity embeddings by a time gate recurrent component and updates the
% relation embeddings by a GRU component.
\label{fig:framework}
\end{figure*}

\section{Related Works}
{\bf Static KG Reasoning.} Existing models for static KG reasoning attempt to
infer missing facts in KGs. Recently, embedding based models~\cite{bordes2013translating,
yang2014embedding, trouillon2016complex, dettmers2018convolutional,
shang2019end} have drawn much attention. As GCN~\cite{kipf2016semi} is a
representative model to combine content and structural features in a graph, some
studies have generalized it to relation-aware GCNs so as to deal with KGs. Among
them, R-GCN~\cite{schlichtkrull2018modeling} extends GCN with relation-specific
filters, and WGCN~\cite{shang2019end} utilizes learnable relation-specific
weights during aggregation. VR-GCN~\cite{ye2019vectorized} and
CompGCN~\cite{vashishth2019composition} jointly embeds both nodes and relations
in a relational graph during GCN aggregation. The above models are all set in
the static KG, and they cannot predict facts in the future.

{\bf Temporal KG Reasoning.} Reasoning over TKG can be categorized into two
settings, interpolation and extrapolation~\cite{jin2020Renet}. For the first
setting, the 
models~\cite{sadeghian2016temporal,esteban2016predicting,garcia2018learning,leblay2018deriving,
dasgupta2018hyte, wu2019efficiently, xu2020temporal, goel2020diachronic,
wu2020temp, han2020dyernie} attempt to infer missing facts at the historical
timestamps. TA-DistMult~\cite{garcia2018learning},
TA-TransE~\cite{garcia2018learning} and TTransE~\cite{leblay2018deriving}
integrate the time when the facts occurred into the embeddings of relations.
HyTE~\cite{dasgupta2018hyte} associates each timestamp with a corresponding
hyperplane. However, they are not able to predict facts at future timestamps and
can not directly compatible with the extrapolation setting.

The extrapolation setting, which this paper focuses on, attempts to predict new
facts at future timestamps based on historical ones. Orthogonal to our work,
some models~\cite{trivedi2017know,trivedi2018dyrep,han2020graph} conduct the
future fact prediction by estimating the conditional probability via a temporal
point process. They are more capable of modeling TKGs with continuous time,
where no facts may occur at the same timestamp. Glean~\cite{deng2020dynamic}
incorporates a word graph constructed by the event summary into the modeling of
future fact prediction. However, not all the events have the summary text in the
practical application. CyGNet~\cite{zhu2020learning} and
RE-NET~\cite{jin2020Renet} are the most related works with us. They attempt to
solve the entity prediction task from the view of each given query, which
encodes the historical facts related to the subject entity in each query. CyGNet
uses a generate-copy network to model the frequency of the historical facts with
the same subject entities and relations to the given queries (i.e., repetitive
patterns). RE-NET uses a GCN and GRU to model the sequence of 1-hop subgraphs
related to the given subject entity. They both neglect the structural
dependencies within KGs at different timestamps and the static properties of
entities. Differently, RE-GCN models the KG sequence as a whole, which considers
all structural dependencies and enables great improvement in efficiency.

\section{Problem Formulation}

\begin{table}
\small
\centering
\setlength{\tabcolsep}{0.3em}
\begin{tabular}{ll}
\toprule
{\bf Notations}  &{\bf Descriptions} \\
$G$, $G^{s}$, $G_{t}$ & TKG, static graph, KG at timestamp $t$ in the TKG \\
$\mathcal{V}$, $\mathcal{R}$, $\mathcal{E}_{t}$    & entity set, relation set, fact set (at $t$) in the TKG\\
$\mathcal{V}^{s}$, $\mathcal{R}^{s}$, $\mathcal{E}^{s}$    & entity set, relation set, edge set in the static graph \\
$\mathbf{H}_{t}$, $\mathbf{R}_{t}$  &embedding matrices of entities, relations at $t$\\
 $\mathbf{H}$,  $\mathbf{R}$   &randomly initialized embedding matrices of entities \\
 &and relations\\
$\mathbf{H}^{s}$ & static embedding matrix of entities in the static graph \\
\bottomrule
\end{tabular}
\caption{Important notations and their descriptions.}
\label{table:denote}
\vspace{-10mm}
\end{table}

A TKG $G$ can be formalized as a sequence of KGs with timestamps, i.e.,
$G=\{G_{1}, G_{2}, ..., G_{t}, ...\}$. Each KG, $G_{t}=(\mathcal{V},
\mathcal{R}, \mathcal{E}_{t})$, at timestamp $t$ is a directed multi-relational
graph, where $\mathcal{V}$ is the set of entities, $\mathcal{R}$ is the set of
relations, and $\mathcal{E}_{t}$ is the set of facts at timestamp $t$ ($t$ is a
discrete integer). Any fact in $\mathcal{E}_{t}$ can be denoted as a quadruple,
$(s, r, o,{t})$, where $s, o \in \mathcal{V}$ and $r\in \mathcal{R}$. It
represents a fact of relation $r$ that occurs with $s$ as its subject entity and
$o$ as its object entity at timestamp $t$. For each quadruple $(s, r, o, t)$,
the inverse quadruple $(o, r^{-1} , s, t)$ is also appended to the dataset. The
static graph is denoted as $G^s = (\mathcal{V}^{s}, \mathcal{R}^s,
\mathcal{E}^s)$, where $\mathcal{V}^{s}$, $\mathcal{R}^{s}$ and
$\mathcal{E}^{s}$ are the entity set, the relation set and the set of edges in
the static graph. The important mathematical notations are in
Table~\ref{table:denote}. 

The entity prediction task aims to predict the missing object entity of a query
$(s, r, ?, t+1)$ and the missing subject entity of a query $(?, r, o, t+1)$. The
relation prediction task attempts to predict the missing relation of a query
$(s, ?, o, t+1)$. Under the assumption that the prediction of the facts at a
future timestamp $t+1$ depends on the KGs at the latest $m$ timestamps
(i.e., $\{G_{t-m+1},...G_{t}\}$) and the information of the historical KG
sequence is modeled in the evolutional embedding matrices of the entities
$\mathbf{H}_{t}\in \mathbb{R}^{|\mathcal{V}| \times d}$ and the relations
$\mathbf{R}_{t} \in \mathbb{R}^{|\mathcal{R}|\times d}$ at timestamp $t$ ($d$ is
the dimension of the embeddings), the problems of the two temporal reasoning
tasks can be formulated as follows:

{\bf Problem 1. Entity Prediction.} Given a query $(s, r, ?, t+1)$, RE-GCN
models the conditional probability vector of all object entities with the
subject entity $s$, the relation $r$ and the historical KG sequence 
$G_{t-m+1:t}$ given:

\begin{equation}
\vec{p}(o|s, r, G_{t-m+1:t}) = \vec{p}(o|s, r, \mathbf{H}_{t}, \mathbf{R}_{t}).
\label{eq:pro1}
\end{equation} 

{\bf Problem 2. Relation Prediction.}  
Given a query $(s, ?, o, t+1)$, RE-GCN models the conditional probability vector
of all relations with the subject entity $s$, the object entity $o$ and the
historical KG sequence $G_{t-m+1:t}$ given:
\begin{equation}
\vec{p}(r|s, o, G_{t-m+1:t}) = \vec{p}(r|s, o,\mathbf{H}_{t}, \mathbf{R}_{t}).
\label{eq:pro2}
\end{equation} 

\section{The RE-GCN Model}
RE-GCN integrates the structural dependencies in a KG at each timestamp, the
informative sequential patterns across temporally adjacent facts, and the static
properties of entities into the evolutional representations of entities and
relations. Based on the learned entity and relation representations, temporal
reasoning at future timestamps can be made with various score functions. Thus
RE-GCN contains an evolution unit and multi-task score functions, as illustrated
in Figure~\ref{fig:framework}. The former is employed to encode the historical
KG sequence and obtain the evolutional representations of entities and
relations. The latter contains score functions for corresponding tasks with the
evolutional representations (i.e., embeddings) at the final timestamp as the
input. 

\subsection{The Evolution Unit}
The evolution unit consists of a relation-aware GCN, two gate recurrent
components, and a static graph constraint component. The relation-aware GCN
attempts to capture the structural dependencies within the KG at each timestamp.
The two gate recurrent components model the historical KG sequence
auto-regressively. Specifically, a time gate recurrent component and a GRU
component get the evolutional representations of entities and relations at each
timestamp correspondingly. The static graph constraint component integrates the
static properties to the evolutional embeddings by adding some constraints
between static embeddings and evolutional embeddings of entities. Formally, the
evolution unit computes a mapping from a sequence of KGs at the latest $m$
timestamps (i.e., $\{G_{t-m+1}, ..., G_{t}\}$) to a sequence of entity embedding
matrices (i.e., $\{\mathbf{H}_{t-m+1},...\mathbf{H}_{t}\}$) and a sequence of
relation embedding matrices (i.e., $\{\mathbf{R}_{t-m+1},..., \mathbf{R}_{t}\}$)
recurrently. Particularly, the input at the first timestamp, including the
entity embedding matrix $\mathbf{H}$ and the relation embedding matrix
$\mathbf{R}$, are randomly initialized. 

\subsubsection{\bf Structural Dependencies among Concurrent Facts.}
The structural dependencies among concurrent facts capture the associations
among the entities through facts and the associations among relations through
the shared entities. Since each KG is a multi-relational graph and GCN is a
powerful model for the graph-structured
data~\cite{schlichtkrull2018modeling,shang2019end,ye2019vectorized,vashishth2019composition},
an $\omega$-layer relation-aware GCN is used to model the structural
dependencies. More specifically, for a KG at timestamp $t$, an object entity $o$
at layer $l\in[0, \omega-1]$ gets information from its subject entities under a
message-passing framework with embeddings of the relations at layer $l$
considered and obtains its embedding at the next $l+1$ layer, i.e., 
\begin{equation}
\vec{h}_{o, t}^{l+1}=
f\left(\frac{1}{c_{o}}\sum_{(s, r), \exists(s, r, o)\in \mathcal{E}_{t} }\mathbf{W}_{1}^{l}(\vec{h}_{s, t}^{l}+ \vec{r}_{t}) +  \mathbf{W}_{2}^{l}\vec{h}_{o,t}^{l}\right), 
\label{eq1}
\end{equation}
where $\vec{h}_{o,t}^{l}$, $\vec{h}_{s,t}^{l}$, $\vec{r}_{t}$ denote the
$l^{th}$ layer embeddings of entities $o, s$ and relation $r$ at timestamp $t$,
respectively; $\mathbf{W}_{1}^{l}$, $\mathbf{W}_{2}^{l}$ are the parameters for
aggregating features and self-loop in the $l^{th}$ layer; $\vec{h}_{s, t}^{l}+
\vec{r}_{t}$ implies the translational property between the subject entity and
the corresponding object entity via the relation $r$; $c_{o}$ is a normalization
constant, equal to the in-degree of entity $o$; $f(\cdot)$ is the RReLU
activation function~\cite{xu2015empirical}. Note that, for those entities that
are not involved in any fact, only a self-loop operation with the extra
parameters $\mathbf{W}_{3}^{l}$ is carried out. Actually, the relation-aware GCN
gets the entity embeddings according to the facts occurred among them at each
timestamp and the self-loop operation can be considered as the self-evolution of
the entities. 

\subsubsection{\bf Sequential Patterns across Temporally Adjacent Facts}
For an entity $o$, the sequential patterns contained in its historical facts
 reflect its behavioral trends and preferences. To cover the historical facts as
 many as possible, the model needs to take all its temporally adjacent facts
 into consideration. As the output of the final layer of the relation-aware GCN,
 $\vec{h}_{o,t-1}^{\omega}$, already models the structure of the adjacent facts
 at timestamp $t-1$, one straightforward and effective approach to contain the
 information of the temporally adjacent facts is to use the output entity
 embedding matrix at $t-1$, $\mathbf{H}_{t-1}$, as the input of the
 relation-aware GCN at $t$, $\mathbf{H}^{0}_{t}$. Therefore, the potential
 sequential patterns are modeled by stacking the $\omega$-layer relation-aware
 GCN. However, although the adjacent KGs are different, the over-smoothing
 problem~\cite{kipf2016semi}, i.e., the embeddings of entities converge to the
 same values, also exists when the repetitive relations occur between the same
 entity pairs at adjacent timestamps~\cite{zhu2020learning}. And when the
 historical KG sequence gets longer, the large number of stacked layers of GCN
 may cause the vanishing gradient problem. Thus,
 following~\cite{li2019deepgcns}, we apply a time gate recurrent component to
 alleviate these problems. In this way, the entity embedding matrix
 $\mathbf{H}_{t}$ is determined by two parts, namely, the output
 $\mathbf{H}^{\omega}_{t}$ of the final layer of the relation-aware GCN at
 timestamp $t$ and $\mathbf{H}_{t-1}$ from the previous timestamp. Formally,
\begin{equation}
\mathbf{H}_{t} =  \mathbf{U}_{t}  \otimes \mathbf{H}^{\omega}_{t} +  (\mathbf{1}-\mathbf{U}_{t})  \otimes \mathbf{H}_{t-1}, \\
\end{equation}
where $\otimes$ denotes the dot product operation. The time gate
$\mathbf{U}_{t}\in\mathbb{R}^{d\times d}$ conducts nonlinear transformation as:
\begin{equation}
\mathbf{U}_{t}  =  \sigma(\mathbf{W}_{4}\mathbf{H}_{t-1} + \mathbf{b}), 
\end{equation}
where $\sigma(\cdot)$ is the sigmoid function and
$\mathbf{W}_{4}\in\mathbb{R}^{d\times d}$ is the weight matrix of the time
gate. Besides, the sequential pattern of relations captures the information of
entities involved in the corresponding facts. Thus, the embeddings of a relation
$\vec{r}_{t}$ at timestamp $t$ are influenced by the evolutional embeddings of
$r$-related entities $\mathcal{V}_{r, t}=\{i|(i, r, o, t)\ or\ (s, r, i, t)\in
\mathcal{E}_{t}\, \}$ at timestamp $t$ and its own embedding at timestamp $t-1$.
Thus, a GRU component is adopted to model the sequential pattern of relations.

By applying mean pooling operation over the embedding matrix of $r$-related
entities at timestamp $t-1$, $\mathbf{H}_{t-1, \mathcal{V}_{r,t}}$, the input of
the GRU at timestamp $t$ for relation $r$, is
\begin{equation}
\vec{r}'_{t}=  [pooling(\mathbf{H}_{t-1, \mathcal{V}_{r,t}}); \vec{r}],
\end{equation}
where $\vec{r}$ is the embedding of relation $r$ in $\mathbf{R}$ and $[;]$
denotes the vector concatenation operation. For the relation that does not have
corresponding facts occurred at timestamp $t$, $\vec{r}'_{t} = \vec{0}$. Then we
update the relation embedding matrix $\mathbf{R}_{t-1}$ to $\mathbf{R}_{t}$ via
the GRU,

\begin{equation}
\mathbf{R}_{t} = GRU(\mathbf{R}_{t-1}, \mathbf{R}'_{t}),
\end{equation}
where $\mathbf{R}'_{t} \in \mathbb{R}^{|\mathcal{R}| \times d}$ consists of
$\vec{r}'_{t}$ of all the relations. Note that, the L2-norm of each line of
$\mathbf{H}_{t}$ and $\mathbf{R}_{t}$ is constrained to $1$.

% \subsection{Static Graph Constraint Unit}
\subsubsection{\bf Static Properties.}
Besides the information contained in the historical KG sequence, some static
properties of entities, which form a static graph, can be seen as the background
knowledge of the TKG and is helpful for the model to learn more accurate
evolutional representations of entities. Thus we incorporate the static graph
into the modeling of the evolutional representations. We construct the static
graphs of the three TKGs from ICEWS based on the entity property information
originally contained in the name strings of entities. Most name strings of
entities therein are in the form of `entity types (country)'. Take an entity
named `Police (Australia)' in ICEWS18~\cite{jin2019recurrent} for example, we
add relation `isA' from this entity to the property entity `Police' and relation
`country' to the property entity `Australia'. The bottom left of
figure~\ref{fig:framework} shows an example of a static graph. Since the static
graph is a multi-relational graph and R-GCN~\cite{schlichtkrull2018modeling} can
model the multi-relational graph without any more extra embeddings for relations.
Thus, we adopt a 1-layer R-GCN~\cite{schlichtkrull2018modeling} without
self-loop to get the static embeddings of entities in the TKG. Then, the update
rule for the static graph is defined as follows:
\begin{equation}
\vec{h}^{s}_i \!=\! \Upsilon\!\left(\frac{1}{c_{i}}\sum_{(r^s, j), \exists(i,r^s,j)) \in \mathcal{E}^s} \mathbf{W}_{r^s}\vec{h}'^{s}_i\left(j\right)\right),
\end{equation}
where $\vec{h}^{s}_i$ and $\vec{h}'^{s}_j$ are the $i^{th}$ and $j^{th}$ lines
of $\mathbf{H}^{s}$ and $\mathbf{H'}^{s}$, which are the output and randomly
initialized input embedding matrices, respectively; $\mathbf{W}_{r^s} \in
\mathbb{R}^{d\times d}$ is the relation matrix of $r^s$ in R-GCN;
$\Upsilon(\cdot)$ is ReLU function; $c_{i}$ is a normalization constant
equal to the number of entities connected with entity $i$. Note that,
$||\vec{h}^{s}_i||_{2}=1$.

To reflect the static properties in the learned sequence of entity embedding
matrices $\mathbf{H}_{t-m}$, $\mathbf{H}_{t-m+1}$,...$\mathbf{H}_{t}$, we
confine the angle between the evolutional embedding and the static embedding of
the same entity not to exceed a timestamp related threshold. It increases over
time since the permitted variable range of the evolutional embeddings of
entities continuously extends over time with more and more facts occurring.
Thus, it is defined as 
\begin{equation}
\theta_{x}=min(\gamma x, 90^{\circ}),
\end{equation}
where $\gamma$ denotes the ascending pace of the angle and $x \in [0,1,.., m]$.
We set the max angle of the two embeddings of an entity to $90^{\circ}$.

Then, the cosine value of the angle between the two embeddings of entity $i$,
denoted as $cos(\vec{h}^{s}_i, \vec{h}_{t-m+x,i})$, should be more than
$cos\theta_{x}$.

%
%\begin{align}
%\theta_{t} &=
%\begin{cases}
%\gamma t&\mbox{if $\theta_{t}\textless90^{\circ}$},\\
%90^{\circ}&\mbox{else},
%\end{cases}
%\end{align}
 %

Thus, the loss of the static graph constraint component at timestamp $t$ can be
defined as below:
\begin{align}
L^{st}_{x} =  \sum_{i=0}^{|\mathcal{V}|-1}\max\{{cos}\theta_{x}-{cos}(\vec{h}^{s}_i, \vec{h}_{t-m+x, i}), 0\}.
\end{align}

The loss of the static graph constraint component is $L^{st} = \sum_{x=0}^{m}L^{st}_{x}$.

\subsection{Score Functions for Different Tasks}
Previous works~\cite{shang2019end, dettmers2018convolutional,
vashishth2019composition} on KG reasoning involve score functions (i.e.,
decoder) to model the conditional probability in Equation (\ref{eq:pro1}) and
(\ref{eq:pro2}), which can be seen as the probability score of candidate triples
$(s, r, o)$. As the previous work~\cite{vashishth2019composition} shows that GCN
with the convolutional score functions gets good performance on KG reasoning and
in order to reflect the translational property of the evolutional embeddings of
entities and relations implied in Equation (\ref{eq1}), we choose
ConvTransE~\cite{shang2019end} as our decoder. ConvTransE contains a
one-dimensional convolution layer and a fully connected layer. We use ConvTransE
($\cdot$) to represent these two layers. Then, the probability vector of all
entities is below: 
\begin{equation}
\vec{p}(o|s, r, \mathbf{H}_{t}, \mathbf{R}_{t}) = \sigma(\mathbf{H}_{t}\text{ConvTransE}(\vec{s}_{t}, \vec{r}_{t})).
\label{eq:sro}
\end{equation} 

Similarly, the probability vector of all the relations is below: 
\begin{equation}
\vec{p}(r|s, o, \mathbf{H}_{t}, \mathbf{R}_{t}) = \sigma(\mathbf{R}_{t}\text{ConvTransE}(\vec{s}_{t}, \vec{o}_{t})),
\end{equation} 
where $\sigma(\cdot)$ is the sigmoid function, $\vec{s}_{t}$, $\vec{r}_{t}$,
$\vec{o}_{t}$ are the embeddings of s, r and o in $\mathbf{H}_{t}$ and
$\mathbf{R_{t}}$, respectively. $\text{ConvTransE}(\vec{s}_{t}, \vec{r}_{t}),
\text{ConvTransE}\\(\vec{s}_{t}, \vec{o}_{t}) \in \mathbb{R}^{d\times1}$. The
details of ConvTransE are omitted for brevity. Note that, ConvTransE can be
replaced by other score functions.

\subsection{Parameter Learning}
Both the entity prediction task and the relation prediction task can be seen as
the multi-label learning problems. Let
$\vec{y}_{t+1}^{e}\in\mathbb{R}^{|\mathcal{V}|}$ and
$\vec{y}_{t+1}^{r}\in\mathbb{R}^{|\mathcal{R}|}$ denote the label vectors for
the two tasks at the timestamp $t+1$, respectively. The elements of vectors
$\vec{y}_{t+1}^{e}\in\mathbb{R}^{|\mathcal{V}|}$ and
$\vec{y}_{t+1}^{r}\in\mathbb{R}^{|\mathcal{R}|}$ are $1$ for facts that do
occur, otherwise, $0$. Then,
\begin{align}
L^{e}= \sum^{T-1}_{t=0} \sum_{(s,r,o,t+1)\in \mathcal{E}_{t+1}}&\sum^{|\mathcal{V}|-1}_{i=0}{y}_{t+1,i}^{e} \log p_{i}(o|s,r, \mathbf{H}_{t},  \mathbf{R}_{t}),\label{eq:L_ent}\\
L^{r} = \sum^{T-1}_{t=0} \sum_{(s,r,o,t+1)\in \mathcal{E}_{t+1}}&\sum^{|\mathcal{R}|-1}_{i=0}{y}_{t+1,i}^{r} \log p_{i}(r|s,o, \mathbf{H}_{t},  \mathbf{R}_{t}),
\end{align}
where $T$ is the number of timestamps in the training set, ${y}_{t+1,i}^{e},
{y}_{t+1,i}^{r}$ is the $i_{th}$ element in $\vec{y}_{t+1}^{e}$,
$\vec{y}_{t+1}^{r}$. $p_{i}(o|s,r, \mathbf{H}_{t}, \mathbf{R}_{t})$ and
$p_{i}(r|s,o, \\\mathbf{H}_{t}, \mathbf{R}_{t})$ are the probability score of
entity $i$ and relation $i$. 

The two temporal reasoning tasks are conducted under the multi-task learning
framework. Therefore, the final loss $L=\lambda_{1}L^{e} + \lambda_{2}L^{r} + L^{st}$.
$\lambda_{1}$ and $\lambda_{2}$ are the parameters that control the loss terms.

\subsection{Computational Complexity Analysis.}
To see the efficiency of the proposed RE-GCN, we analyze the computational
complexity of its evolution unit. The time complexity of the relation-aware
GCN at a timestamp $t$ is ${O}(|\mathcal{E}|\omega)$, where $|\mathcal{E}|$ is
the maximum number of concurrent facts in the historical KG sequence. The
pooling operation to get the input of the GRU component at every timestamp has
the time complexity $O(|\mathcal{R}|D)$, where $D$ is the maximum number of
entities involved in a relation at timestamp $t$ and $|\mathcal{R}|$ is the size
of relation set. The time complexity to get the static embeddings is
$O(|\mathcal{E}^{s}|)$. As we unroll $m$ steps for the GRU component and the
relation-aware GCN, the time complexity for the evolution unit is finally
${O}(m(|\mathcal{E}|\omega+|\mathcal{R}|D)+|\mathcal{E}^{s}|)$.

\begin{table*}
  \centering
  \begin{tabular}{lrrrrrrrr}
  \toprule
  Datasets  & $|\mathcal{V}|$  & $|\mathcal{R}|$  & $|\mathcal{E}_{train}|$ &
  $|\mathcal{E}_{valid}|$ &$|\mathcal{E}_{test}|$   &$|\mathcal{E}^{s}|$
  &$|\mathcal{V}^{s}|$  & Time interval\\
  \midrule
  ICEWS18       &23,033   &256     &373,018   &45,995  &49545  &29,774 &8,647  &24 hours\\
  ICEWS14       &6,869    &230     &74,845    &8,514   &7,371   &8,442  &3,499  &24 hours    \\
  ICEWS05-15    &10,094   &251     &368,868   &46,302  &46,159  &12,392 &5,179  &24 hours  \\
  WIKI          &12,554   &24      &539,286   &67,538  &63,110  &--    &--    &1 year\\
  YAGO          &10,623   &10      &161,540   &19,523  &20,026  &--    &--    &1 year \\
  GDELT         &7,691    &240     &1,734,399  &238,765 &305,241 &--    &--    &15 mins\\
  %  &29774    &8442    &12392\\
  %   &8647    &3499    &5179\\
  \bottomrule
  \end{tabular}
  \caption{Statistics of the datasets ($|\mathcal{E}_{train}|$, $|\mathcal{E}_{valid}|$, $|\mathcal{E}_{test}|$ are the numbers of facts in training, validation, and test sets.).}
  \label{table:datasets}
  \vspace{-5mm}
  \end{table*}

\section{Experiments}
\subsection{Experimental Setup}
\subsubsection{\bf Datasets} There are six typical TKGs commonly used in
previous works, namely, ICEWS18~\cite{jin2020Renet},
ICEWS14~\cite{garcia2018learning}, ICEWS05-15~\cite{garcia2018learning},
WIKI~\cite{leblay2018deriving}, YAGO~\cite{mahdisoltani2014yago3} and
GDELT~\cite{leetaru2013gdelt}. The first three datasets are from the Integrated
Crisis Early Warning System~\cite{boschee2015icews} (ICEWS).
GDELT~\cite{jin2020Renet} is from the Global Database of Events, Language, and Tone
~\cite{leetaru2013gdelt}. We evaluate RE-GCN on all these datasets. We divide
ICEWS14 and ICEWS05-15 into training, validation, and test sets, with a
proportion of 80\%, 10\% and 10\% by timestamps following ~\cite{jin2020Renet}.
The details of the datasets are presented in Table~\ref{table:datasets}. The
time interval represents time granularity between temporally adjacent facts.

\subsubsection{\bf Evaluation Metrics} 

In the experiments, $MRR$ and $Hits@\{1\\,3,10\}$ are employed as the metrics
for entity prediction and relation prediction. For the entity prediction task on
WIKI and YAGO, we only report the $MRR$ and $Hits@3$ results because the results
of $Hits@1$ were not reported by the prior work RE-NET~\cite{jin2020Renet}.

As mentioned in ~\cite{han2020graph, ding2021temporal, jain2020temporal}, the
filtered setting used
in~\cite{bordes2013translating,jin2020Renet,zhu2020learning}, which removes all
the valid facts that appear in the training, validation, or test sets from the
ranking list of corrupted facts, is not suitable for temporal reasoning tasks.
Take a typical query $(s, r, ?, t_{1})$ with answer $o_{1}$ in the test set for
example, and assume there is another fact $(s, r, o_{2}, t_{2})$. Under this
filtered setting, $o_{2}$ will be wrongly considered a correct answer and thus
removed from the ranking list of candidate answers. However, $o_{2}$ is
incorrect for the given query, as $(s, r, o_{2})$ occurs at timestamp $t_{2}$
instead of $t_{1}$. Thus, the filtered setting may probably get incorrect higher
ranking scores. Without loss of generality, only the experimental results under
the raw setting are reported.

\begin{table*}[htb]
\centering
\begin{tabular}{lrrrrrrrrrrrr}
\toprule
\multirow{2}{*}{Model} &\multicolumn{4}{c}{ICE18} &\multicolumn{4}{c}{ICE14} &\multicolumn{4}{c}{ICE05-15}\\
\cmidrule(r){2-5}  \cmidrule(r){6-9} \cmidrule(r){10-13}&MRR &H@1 &H@3 &H@10 &MRR &H@1 &H@3 &H@10 &MRR &H@1 &H@3 &H@10\\
\midrule
% TransE\!\!\!       & 12.37   & 1.51   & 15.99   & 34.65     &17.46    &1.39    &21.34    &45.49      &18.80    &2.60    &29.06    &49.03 \\
DistMult\!\!\!     & 13.86   & 5.61   & 15.22   & 31.26       &20.32    &6.13    &27.59    &46.61     &19.91    &5.63    &27.22    &47.33        \\
ComplEx\!\!\!   & 15.45   & 8.04   & 17.19  & 30.73     &22.61    &9.88    &28.93    &47.57     &20.26    &6.66    &26.43    &47.31    \\
R-GCN\!\!\!      & 15.05  & 8.13  &16.49  & 29.00	&28.03    &19.42    &31.95    &44.83     &27.13    &18.83    &30.41    &43.16        \\
% VR-GCN\!\!\!    & 18.03  &10.32  &19.70    &33.93       &28.70    &20.38    &31.94    &44.82       &27.58    &19.36    &30.72    &44.01       \\
% SCAN\!\!\!         &21.46 &11.62 &24.57 &41.69    &32.00  &22.68  &36.12 &50.51      &31.18 &20.48 &35.25 &52.86  \\
ConvE\!\!\!        & 22.81  &13.63  &25.83  &41.43      &30.30   &21.30 &34.42  &47.89       &31.40 &21.56  &35.70  &50.96   \\
ConvTransE\!\!\!  &23.22 &14.26 &26.13 &41.34        &31.50  &22.46  &34.98  &50.03       &30.28 &20.79  &33.80  &49.95\\
RotatE\!\!\!  &14.53   &6.47 &15.78  &31.86        &25.71 &16.41  &29.01  &45.16  &19.01  &10.42  &21.35  &36.92\\
\midrule
HyTE\!\!\!             & 7.41  & 3.10  & 7.33  & 16.01      &16.78    &2.13  &24.84    & 43.94    &16.05 &6.53 &20.20 & 34.72\\
TTransE\!\!\!        & 8.44  & 1.85  & 8.95  & 22.38      &12.86  &3.14  &15.72  &33.65                 &16.53    &5.51    &20.77    &39.26\\
TA-DistMult\!\!\!   & 16.42  & 8.60   & 18.13  & 32.51   &26.22  &16.83  &29.72
&45.23              &27.51    &17.57    &31.46    &47.32\\
\midrule
% Know-Evolve+MLP &7.41 &3.31 &7.87 &14.76  &-- &-- &-- &-- &-- &-- &-- &--\\
% DyRep+MLP     &7.82 &3.57 &7.73 &16.33  &-- &-- &-- &-- &-- &-- &-- &--\\
RGCRN          &23.46  &14.24  &26.62  &41.96 &33.31 &24.08 &36.55 &51.54 &35.93 &26.23 &40.02 &54.63\\
% \midrule
%\scriptsize{KnowEvolve(ms)}\!\!\!    &9.29 &5.11 &9.62 &17.18 &-- &-- &-- &-- &-- &-- &-- &--\\
% GCRN(ms)\!\!\!     &21.14    &12.51    &23.60    &38.37    &34.32    &24.75    &38.13    &53.01    &28.50    &18.78    &32.13    &48.39\\
% EvolveGCN(ms)\!\!\!    &21.04    &13.15    &23.16    &36.59    &33.27    &24.00    &36.75    &51.67    &25.79 &16.73	&28.72	&44.67\\
CyGNet            &24.98  &15.54  &28.58  &43.54  &34.68  &25.35  &38.88  &53.16
&35.46  &25.44  &40.20  &54.47\\
RE-NET\!\!\!     & 26.17  & 16.43  & 29.89  & 44.37  & 35.77  & 25.99  & 40.10  & 54.87  & 36.86  & 26.24  & 41.85  & 57.60\\
RE-GCN\!\!\!   &\textbf{27.51}  & \textbf{17.82}  & \textbf{31.17}  & \textbf{46.55}   & \textbf{37.78}  & \textbf{27.17}  & \textbf{42.50}  & \textbf{58.84}  & \textbf{38.27}  &\textbf{27.43}  & \textbf{43.06}  & \textbf{59.93}\\
\midrule
% \midrule
% GCRN(ss)\!\!\!         &24.85  & 15.40  & 27.90  & 43.79 & 36.95    &27.00    &41.50    &56.12    &35.01    &24.69    &39.64    &55.68\\
% EvolveGCN(ss)\!\!\! &24.93    &15.61    &27.81    &43.93    &36.35    &26.37    &40.73    &56.09	&32.45	&22.57	&36.68	&52.11\\
% RE-NET w.GT\!\!\!   & 27.87   & 18.12   & 31.60   & 46.65  & 39.13  & 29.53  & 43.62  & 57.70  & 42.92  & 32.65  & 48.40  & 62.01\\
RE-GCN w. GT\!\!\! & \textbf{30.55}  & \textbf{20.00}  & \textbf{34.73}  & \textbf{51.46}  & \textbf{41.50}  & \textbf{30.86}  & \textbf{46.60}  & \textbf{62.47}  & \textbf{46.41}  & \textbf{35.17}  & \textbf{52.76}  & \textbf{67.64}\\
\bottomrule
\end{tabular}
\caption{Performance (in percentage) for the entity prediction task on ICEWS18, ICESW14 and ICEWS05-15 with raw metrics.}
\label{table:ea1}
\vspace{-4mm}
\end{table*}

\begin{table*}[htb]
  \centering
  \begin{tabular}{lrrrrrrrrrr}
  \toprule
  \multirow{2}{*}{Model} &\multicolumn{3}{c}{WIKI} &\multicolumn{3}{c}{YAGO} &\multicolumn{4}{c}{GDELT}\\
  \cmidrule(r){2-4}  \cmidrule(r){5-7} \cmidrule(r){8-11} &MRR &H@3 &H@10 &MRR &H@3 &H@10  &MRR  &H@1 &H@3 &H@10\\
  \midrule
  % TransE\!\!\!       & 26.21   & 31.25   & 39.06   & 33.85     &48.19    &59.50 &7.84  &0.00 &8.92  &23.30\\
  DistMult\!\!\!     & 27.96   & 32.45   & 39.51   & 44.05       &49.70  &59.94  &8.61  &3.91 &8.27 &17.04          \\
  ComplEx\!\!\!   & 27.69   & 31.99   & 38.61  & 44.09     &49.57    &59.64
  &9.84 &5.17 &9.58 &18.23 \\
  R-GCN\!\!\!      & 13.96  & 15.75  &22.05  & 20.25	&24.01    &37.30    &12.17
  &7.40 &12.37  &20.63 \\
  % VR-GCN\!\!\!     \\
  % SCAN\!\!\!         \\
  ConvE\!\!\!        & 26.03  &30.51  &39.18  &41.22      &47.03   &59.90 &18.37  &11.29  &19.36  &32.13\\
  ConvTransE\!\!\!   &30.89 &34.30  &41.45  &46.67  &52,22  &62,52  &19.07
  &11.85  &20.32  &33.14\\
  RotatE    & 26.08 &31.63  &38.51  &42.08  &46.77  &59.39  &3.62 &0.52 &2.26 &8.37\\
  \midrule
  HyTE\!\!\!             & 25.40  & 29.16  & 37.54  & 14.42      &39.73 &46.98 &6.69 &0.01  &7.57 &19.06\\
  TTransE\!\!\!        & 20.66  & 23.88  & 33.04  & 26.10      &36.28  &47.73 &5.53  &0.46 &4.97  &15.37\\
  TA-DistMult\!\!\!   & 26.44  & 31.36   & 38.97  & 44.98   &50.64  &61.11 &10.34  &4.44 &10.44  &21.63\\
  \midrule
  % Know-Evolve+MLP\!\!\! &10.54  &13.08  &20.21  &5.23 &5.63 &10.23  &15.88  &11.66  &15.69  &22.28\\
  % DyRep+MLP\!\!\!   &10.41  &12.06  &20.93 &4.98  &5.54 &10.19  &16.25  &11.78  &16.45  &23.86\\
  RGCRN\!\!\!  &28.68  &31.44  &38.58  &43.71  &48.53  &56.98  &18.63  &11.53  &19.80  &32.42\\
  % \midrule
  %\scriptsize{KnowEvolve(ms)}\!\!\!    &9.29 &5.11 &9.62 &17.18 &-- &-- &-- &-- &-- &-- &-- &--\\
  % GCRN(ms)\!\!\!     &28.68    &31.44    &38.58    &43.71    &48.53    &56.98   \\
  % EvolveGCN(ms)\!\!\!      \\
  CyGNet\!\!\!    &30.77   &33.83  &41.19  &46.72   &52.48  &61.52  &18.05  &11.13  &19.11  &31.50\\
  RE-NET\!\!\!     & 30.87  & 33.55  & 41.27  & 46.81  & 52.71  & 61.93  &\textbf{19.60} &\textbf{12.03}  &\textbf{20.56}  &\textbf{33.89}\\
  RE-GCN\!\!\!   &\textbf{39.84}  & \textbf{44.43}  & \textbf{53.88}  & \textbf{58.27}   & \textbf{65.62}  & \textbf{75.94} &19.15  &11.92  &20.40  &33.19\\
  \midrule
  % GCRN(ss)\!\!\!          \\
  % EvolveGCN(ss)\!\!\!   \\
  % RE-NET w.GT\!\!\!   & 32.44  &35.42  &43.16  &48.60  &54.20  &63.59 &21.29  &13.99  &22.53  &35.59\\
  RE-GCN w. GT\!\!\! & \textbf{51.53}  & \textbf{58.29}  & \textbf{69.53}  & \textbf{63.07}  & \textbf{71.17}  & \textbf{82.07} &19.31  &11.99  &20.61  &33.59\\
  \bottomrule
  \end{tabular}
  \caption{Performance (in percentage) for the entity prediction task on WIKI,
  YAGO and GDELT with raw metrics.}
  \label{table:ea2}
  \vspace{-4mm}
  \end{table*}

\begin{table}
  \centering
  \setlength{\tabcolsep}{0.3em}
  \begin{tabular}{lrrrrrr}
  \toprule
  Model  & ICE18  & ICE14  & ICE05-15  &WIKI & YAGO &GDELT\\
  % \midrule
  ConvE    & 37.73    & 38.80   &37.89  &78.23  &91.33  &18.84\\
  ConvTransE   & 38.00  &38.40 &38.26 &86.64  &90.98  &18.97 \\
  % \midrule
  RGCRN   & 37.14		&38.04	&38.37 &88.88 &90.18  &18.58\\
  % EvolveGCN(ms)   &36.40			&37.41	&35.76\\
  RE-GCN    &\textbf{39.48}  &  \textbf{39.73}    & \textbf{38.56}  & \textbf{95.63} & \textbf{95.18}  &\textbf{19.17} \\
  % \midrule
  \midrule
  % GCRN(ss)        		& 38.07/42.72		&38.28/43.18	&39.33/44.33\\
  % EvolveGCN(ss)       	&37.48/41.93			&38.33/43.05	&35.74/40.39 \\
  RE-GCN w.GT    &\textbf{40.53}  & \textbf{41.06}   & \textbf{40.63} & \textbf{97.92} & \textbf{97.74}  &\textbf{19.22}    \\
  \bottomrule
  \end{tabular}
  \caption{Performance on the relation prediction task.}
  \label{table:et}
  \vspace{-9mm}
  \end{table}

\subsubsection{\bf Baselines}  The RE-GCN model is compared with two categories
of models: static KG reasoning models and TKG reasoning models.
DistMult~\cite{yang2014embedding}, ComplEx~\cite{trouillon2016complex},
R-GCN~\cite{schlichtkrull2018modeling}, ConvE~\cite{dettmers2018convolutional},
ConvTransE~\cite{shang2019end}, RotatE~\cite{sun2018rotate} are selected as
static models. HyTE~\cite{dasgupta2018hyte}, TTransE~\cite{leblay2018deriving}
and TA-DistMult~\cite{garcia2018learning} are selected as the temporal models
under the interpolation setting. For temporal models under the extrapolation
setting,
% Know-Evolve+MLP~\cite{trivedi2017know}, DyRep+MLP~\cite{trivedi2018dyrep},
CyGNet~\cite{zhu2020learning} and RE-NET~\cite{jin2020Renet} are compared. For
Know-evolve and DyRep, RE-NET extends them to temporal reasoning task but does not
release their codes. Thus, we only report the results in their papers.
Besides, GCRN~\cite{seo2018structured} is the model for homogeneous graphs and
RE-NET extends it to R-GCRN by replacing GCN with R-GCN. 

\subsubsection{\bf Implementation Details} For the evolution unit, the embedding
dimension $d$ is set to 200; the number of layers $\omega$ of the relation-aware
GCN is set to 1 for YAGO and 2 for the other datasets; the dropout rate is set
to 0.2 for each layer of the relation-aware GCN. We perform grid search on the
length of the historical graph sequence (1, 15) and the ascending pace of the
angle $\gamma$ ($1^{\circ}$-$20^{\circ}$) on the validation sets. The optimal
lengths of history $m$ for ICEWS18, ICEWS14, ICEWS05-15, WIKI, YAGO, and GDELT
are 6, 3, 10, 2, 1, 1, respectively. $\gamma$ is experimentally set to
$10^{\circ}$. Adam~\cite{kingma2014adam} is adopted for parameter learning with
the learning rate 0.001. As for the R-GCN used in the static graph constraint
component, we set the block dimension to $2\times2$ and the dropout rate for
each layer to 0.2. For ConvTransE, the number of kernels is set to 50, the
kernel size is set to $2\times3$ and the dropout rate is set to $0.2$. For the
joint learning of the entity prediction task and the relation prediction task,
$\lambda_{1}$ and $\lambda_{2}$ are experimentally set to $0.7$ and $0.3$,
respectively. The statistics of the static graphs are presented in
Table~\ref{table:datasets}. We only report the results without the static graph
constraint on WIKI, YAGO and GDELT because the static information is missing in
these datasets. To conduct the multi-step inference~\cite{jin2020Renet} in the
validation set and the test set, we evaluate the performance of RE-GCN with the
evolutional embeddings of entities and relations at the final timestamp of the
training set as the input of score functions following~\cite{zhu2020learning}.
Besides, we also report the results of the models with ground truth history
given during multi-step inference on the test set, namely, w. GT. All
experiments are carried out on Tesla V100. Codes are avaliable at
\href{https://github.com/Lee-zix/RE-GCN}{https://github.com/Lee-zix/RE-GCN}.
% We release the Pytorch implementation of RE-GCN on
% \href{https://github.com/Lee-zix/RE-GCN}{the authors' webpage}.

\subsection{Experimental Results}

\subsubsection{\bf Results on Entity Prediction}  \label{result:ee} The
experimental results on the entity prediction task are presented in
Tables~\ref{table:ea1} and \ref{table:ea2}. RE-GCN consistently outperforms the
baselines on the three ICEWS datasets, WIKI and YAGO. The results convincingly
verify its effectiveness. Specifically, RE-GCN significantly outperforms the
static models (i.e., those in the first blocks of Tables~\ref{table:ea1} and
~\ref{table:ea2}) because RE-GCN considers the sequential patterns across
timestamps. RE-GCN performs better than the temporal models for the
interpolation setting (i.e., those in the second blocks of
Tables~\ref{table:ea1} and ~\ref{table:ea2}) because RE-GCN additionally
captures temporally sequential patterns and static properties of entities. It
can thus obtain more accurate evolutional representations for the unobserved
timestamps. Especially, RE-GCN outperforms the temporal models for the
extrapolation setting (i.e., those in the third blocks of Tables~\ref{table:ea1}
and ~\ref{table:ea2}). It outperforms RGCRN because the newly designed graph
convolution operation and the two recurrent components in the evolution unit
learn better evolutional embeddings and the static graph helps learn better
evolutional embeddings of entities. CyGNet and RE-NET's good performance
verifies the importance of the repetitive patterns and 1-hop neighbors to the
entity prediction task. Despite this, it is not surprising that RE-GCN performs
better than CyGNet because there is much useful information except the
repetitive patterns in the history. RE-GCN also performs better than RE-NET,
which neglects the structural dependencies within a KG and the static properties
of entities. By capturing more comprehensive structural dependencies and
sequential patterns, RE-GCN outperforms RE-NET on most datasets.  From the last
two lines in Tables~\ref{table:ea1} and ~\ref{table:ea2}, it can be observed
that the performance gap between the last two lines becomes large when the time
interval between two adjacent timestamps of the datasets becomes large. For the
two datasets, WIKI and YAGO, with the time interval as one year, the model's
performance drops rapidly without knowing the ground truth history. This is
because the evolutional representations become inaccurate when the time interval
is large during the multi-step inference.

Note that RE-GCN even achieves the improvements of 8.97/11.46\% in MRR,
10.60/12.91\% in Hits@3 and 12.61/14.01\% in Hits@10 over the best baseline on
WIKI and YAGO. For the two datasets, there are more structural dependencies
within the KG at each timestamp because the time interval is much large than the
other datasets. Therefore, only modeling repetitive patterns or one-hop neighbors
will loses a lot of structural dependencies and sequential patterns. The results
demonstrate that RE-GCN is more capable of modeling these datasets containing
complex structural dependencies among concurrent facts.

The experimental results of static models and temporal models are similarly poor
on GDELT, as compared with those of the other five datasets. We further analyze
the GDELT dataset and find that many of its entities are abstract concepts that
do not indicate specific entities (e.g., POLICE and GOVERNMENT). Among the top
50 frequent entities, 28 are abstract concepts and 43.72\% corresponding facts
involve abstract concepts. Those abstract concepts make the temporal reasoning
for some entities under the raw setting almost impossible, since we cannot
predict a government's activities without knowing which country it belongs to.
Thus, all the models can only predict partial facts in the GDELT dataset and get
similar results. Besides, the noise produced by the abstract concepts influences
the evolutional representations of other entities as RE-GCN models the KG
sequence as a whole, which makes the results of RE-GCN a little worse than
RE-NET.

\subsubsection{\bf Results on Relation Prediction}
Since some models are not designed for the relation prediction task and for
space limitation, we select the typical ones from the baselines and present the
experimental results in terms of only MRR in Table~\ref{table:et}. In more
detail, we select ConvE~\cite{dettmers2018convolutional},
ConvTransE~\cite{shang2019end} from the static models, as well as
RGCRN~\cite{seo2018structured} from the temporal models. RE-NET and CyGNet are
not adopted, as they cannot be applied to the relation prediction task directly.
It can observe that RE-GCN performs better than all the baselines. The
outperformance of RE-GCN demonstrates that our evolution unit can obtain more
accurate evolutional representations by modeling the history comprehensively.

The performance gap between RE-GCN and other baselines on the relation
prediction task is smaller than the entity prediction task. It is because the
number of relations is much less than the number of entities. Fewer candidates
make the relation prediction task much easier than the entity prediction task.
The performance on WIKI and YAGO is much better than the other datasets because
the numbers of relations in the two datasets are only 24 and 10, respectively.
The results on the GDELT dataset for the static models and the temporal models
are also similarly poor, which verifies our observations mentioned in
Section~\ref{result:ee} again.

\subsection{Comparison on Prediction Time}
\label{sec:time}

\begin{figure}[htbp]
  \centering
  \includegraphics[width=3.4in]{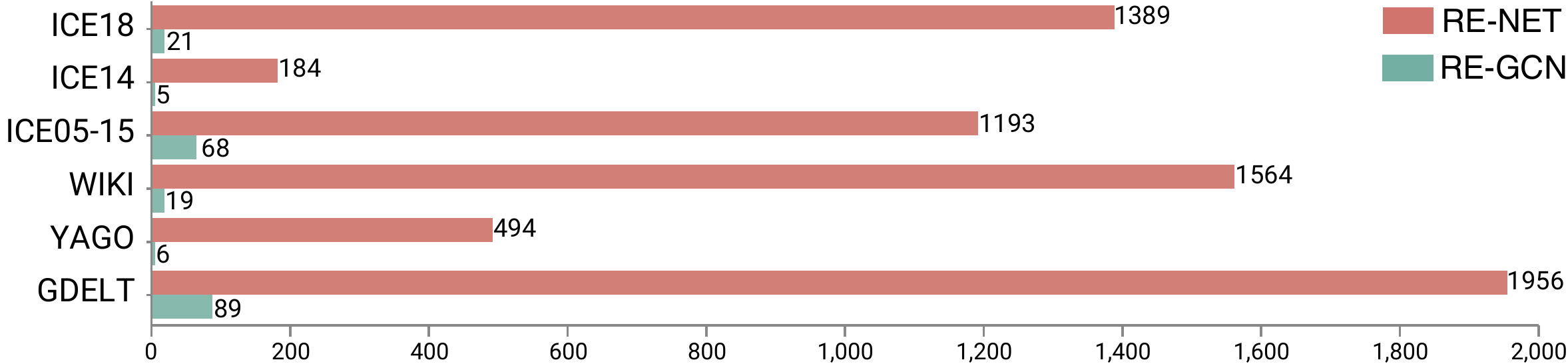}
  \caption{Runtime (seconds) comparison to RE-NET.}
  \label{fig_time}
  \vspace{-3mm}
  \end{figure}

To investigate the efficiency of RE-GCN, we compare RE-GCN to RE-NET in terms of
runtime for entity prediction on the test set under the same environment. For a
fair comparison, the two models conduct entity prediction with the ground truth
history given. From the results in Fig~\ref{fig_time}, it can be seen that
RE-GCN is much faster than RE-NET by 66, 36, 17, 82, 82, 22 times on ICE18,
ICE14, ICE05-15, WIKI, YAGO, and GDELT, respectively. It is because RE-NET
processes individual queries one by one for each timestamp, while RE-GCN
characterizes the evolutional representation learning from the view of KG
sequence and obtains the representations for all the queries at the same
timestamp simultaneously. Therefore, RE-GCN is more efficient than the best
baseline, RE-NET.

\begin{table}
  \centering
  \setlength{\tabcolsep}{0.3em}
  \begin{tabular}{lrrrrrr}
  \toprule
  Model  & ICE18  & ICE14  & ICE05-15 &WIKI & YAGO &GDELT\\
  \midrule
  RE-GCN w. GT & \textbf{30.55}   & \textbf{41.50}  & \textbf{46.41} &51.53
  &\textbf{63.07}  &\textbf{19.31}\\
  RE-NET w. GT & 27.87   &39.13  &42.92  &32.44  &48.60  &21.29\\
  \midrule
  -EE w. GT & 23.22   &31.50  &30.28  &30.89  &46.67  &19.07\\
  +FCN w. GT  &29.32    &40.34  &45.89   &46.00  &58.96 &19.02\\
  % +RGCN    &29.63    &39.13    &43.82\\
  -st w. GT &29.10  &39.48   &44.68 &--  &--   &--\\
  % ER--$L_{evt}$)\!\!\! & 30.38 &41.05 &45.92 \\
  -tg w. GT    &24.51  &34.85 &37.65 &\textbf{51.70} &62.23  &18.55\\
  \bottomrule
  \end{tabular}
  \caption{Ablation studies on entity prediction.}
  \label{table:ablation_ent}
  \vspace{-7mm}
  \end{table}

  \begin{table}
  \centering
  \setlength{\tabcolsep}{0.3em}
  \begin{tabular}{lrrrrrr}
  \toprule
  Model  & ICE18  & ICE14  & ICE05-15 &WIKI &YAGO &GDELT\\
  \midrule
  RE-GCN w. GT   &\textbf{40.53}	 &\textbf{41.06}    &\textbf{40.63}
  &\textbf{97.92} &\textbf{97.74}  &\textbf{19.22}     \\
  RGCRN w. GT &38.07 &38.28 &39.33 &90.12 &91.27  &18.73\\
  \midrule
  -EE w. GT     & 38.00  &38.40 &38.26 &86.64  &90.98  &18.97 \\
  +FCN w. GT   &39.63    &40.23    &40.55   &97.23  &93.66 &19.03\\
  % +RGCN    &{38.12}    &{38.48}    &39.10\\
  -st w. GT   &39.23    &40.00 &40.38 &--  &--   &--\\
  % RE-GCN(--$L_{ent}$)   &40.26  &40.32 &\textbf{41.15}      \\
  -tg w. GT      &37.47  &38.14 &37.62  &97.56  &93.86  &18.94\\
  \bottomrule
  \end{tabular}
  \caption{Ablation studies on relation prediction.}
  \label{table:ablation_evt}
  \vspace{-7mm}
  \end{table}

  \begin{table*}[htb]
    \centering
    \begin{tabular}{rrlc}
    \toprule
    % & &Query &Answer\\
    \makecell[l]{History at $t-2$} &\makecell[l]{History at $t-1$}
    &\makecell[l]{Query at $t$} &Answer\\
    \midrule
    \makecell[l]{\textbf{New Zealand},\\ Host a visit, \textbf{\textit{FIJI}}} &\makecell[l]{\textbf{New Zealand}, \\Criticize, Japan}
    &\makecell[l]{\textbf{New
    Zealand},\\Diplomatic cooperation, ?} &\makecell[c]{\textbf{\textit{FIJI}}}\\
    
    \midrule
    \makecell[l]{\textbf{\textit{Citizen}}, \\Conduct bombing, Government} &\makecell[l]{Government, \\Make statement, \textbf{Defense
    ministry} \\\textbf{Defense Ministry}, \\Make Statement, Police}  &\makecell[l]{\textbf{Defense ministry},\\ Make
    request, ?}  &\makecell[c]{\textbf{\textit{Citizen}}}\\
    
    \midrule
    \makecell[l]{\textbf{Protester}, \\Demonstrate, Defense ministry} &\makecell[l]{Defense
    ministry, \\Endorse, \textbf{Police}} &\makecell[l]{\textbf{Protester}, ?, \textbf{Police}} &\makecell[c]{\textbf{\textit{Protest violently}}} \\
    % \midrule
    % \makecell[l]{t-2} &\makecell[l]{t-1} &\makecell[l]{t} & \\
    % \midrule
    
    %  & History  & ICE14  & ICE05-15 &WIKI &YAGO &GDELT\\
    % \midrule
    % RE-GCN   &\textbf{40.53}	 &\textbf{41.06}    &40.63   &93.78 &97.74  &19.22     \\
    % -EE     & 38.00  &38.40 &38.26 &86.64  &90.98  &18.97 \\
    % +FCN   &39.63    &40.23    &40.55   &97.23  &93.66 &19.03\\
    % % +RGCN    &{38.12}    &{38.48}    &39.10\\
    % -st   &39.23    &40.00 &40.38 &--  &--   &--\\
    % % RE-GCN(--$L_{ent}$)   &40.26  &40.32 &\textbf{41.15}      \\
    % -tg      &37.47  &38.14 &37.62  &97.94  &93.86\\
    \bottomrule
    \end{tabular}
    \caption{Case study. The first two lines are two cases for entity prediction and the last line is a case for relation prediction.}
    \label{table:case}
    \vspace{-3mm}
    \end{table*}
    
    %--------------------------------------------------------------------------------------------------------------
    \begin{table*}[htb]
      \centering
      \begin{tabular}{ccccccccccccccccccccc}
      \toprule 
      \multirow{1}{*}{Tasks} &\multicolumn{10}{c}{Entity Prediction} &\multicolumn{10}{c}{Relation Prediction} \\
      % \cline{2-21}
      \cmidrule(r){0-11}  \cmidrule(r){12-21}% &\cmidrule(r){8-11}
      \multirow{2}{*}{Subsets} & \multicolumn{5}{c}{seen} &\multicolumn{5}{c}{unseen} &\multicolumn{5}{c}{seen} &\multicolumn{5}{c}{unseen} \\
      \cmidrule(r){2-6} \cmidrule(r){7-11} \cmidrule(r){12-16}  \cmidrule(r){17-21}
      % &0 &1 &2$\sim$3 &4$\sim$10 &>10  &0 &1 &2$\sim$3 &4$\sim$10 &>10
      % &0 &1 &2$\sim$3 &4$\sim$10 &>10 &0 &1 &2$\sim$3 &4$\sim$10 &>10 \\
      &a &b &c &d &e  &a &b &c &d &e 
      &a &b &c &d &e  &a &b &c &d &e \\
      \hline
      % &  & unseen  & seen & unseen\\
      % \cmidrule(r){2-2}  \cmidrule(r){4-5}  \cmidrule(r){6-7} \cmidrule(r){8-9}
      % \cmidrule(r){10-11}  \cmidrule(r){12-13} \cmidrule(r){14-15}
      % \cmidrule(r){16-17} \cmidrule(r){18-19}  \cmidrule(r){20-21}
    
      % &0 &1 &2$\sim$3 &4$\sim$10 &>10  &0 &1 &2$\sim$3 &4$\sim$10 &>10
      % &0 &1 &2$\sim$3 &4$\sim$10 &>10 &0 &1 &2$\sim$3 &4$\sim$10 &>10 \\
      \% &0 & 1.8 &3.9 &13.8  &21.0 &19.8 &7.6  &9.1  &9.8  &13.2           &0 &22.3  &6.0 &5.2  &1.0 &37.3 &8.7  &11.4  &5.2  &1.0 \\
      \hline
      H@3  &0 &78.6  &78.4  &71.6  &53.3  &35.2  &33.2  &28.8  &21.4  &6.9  &0 &76.1 &68.8  &51.1  &29.0  &48.6  &42.2  &37.0  &33.5  &30.6\\
    
      \bottomrule
      \end{tabular}
      \caption{Hits@3 on different subsets from the validation set of
      ICEWS18. The \% row shows the proportion of each subset. }
      \label{table:detail analysis}
      \vspace{-5mm}
      \end{table*}
    %--------------------------------------------------------------------------------------------------------------

\subsection{Ablation Studies}
To eliminate the bias between training and testing on the results, we conduct
all ablation studies with ground truth history given on the test sets. To
further show the effectiveness of each part of RE-GCN, we also report the
results of RE-NET w. GT in Table~\ref{table:ablation_ent} and the results of
RGCRN w. GT in Table~\ref{table:ablation_evt}.

\subsubsection{\bf Impact of the Evolution Unit} 
To demonstrate how the evolution unit contributes to the final results of
RE-GCN, we conduct experiments of only using the ConvTransE score function with
the randomly initialized learnable embeddings. The results denoted as -EE w. GT, are
demonstrated in Tables~\ref{table:ablation_ent} and ~\ref{table:ablation_evt}.
It can be observed that removing the evolution unit has a great impact on the
results for all the datasets except GDELT, suggesting that modeling the
historical information is vital for all the datasets. For GDELT, only using the
ConvTransE can get good results. It also matches our observations mentioned in
Section~\ref{result:ee}.

To further verify the effectiveness of our evolution unit under different score
functions, we replace the ConvTransE in RE-GCN with a simple one layer Fully
Connected Network (FCN), denoted as +FCN w. GT. The experimental results are
presented in Tables~\ref{table:ablation_ent} and ~\ref{table:ablation_evt}. It
can be observed that the results are worse than RE-GCN w. GT on most datasets.
It matches the observation in ~\cite{ye2019vectorized}, the convolutional score
functions are more suitable for the GCN. However, even with a simple score
function, +FCN w. GT still shows strong performance on both entity and relation
predictions.
% That
% means, , our model still outperforms all the
% baselines and the evolution unit is effective. 

\subsubsection{\bf Impact of the Static Graph Constraint Component}  
The results denoted as --st w. GT in Tables~\ref{table:ablation_ent} and
\ref{table:ablation_evt} demonstrate the performance of RE-GCN without the
static graph constraint component. It can be seen that --st w. GT performs
consistently worse than RE-GCN w. GT in ICEWS datasets, which justifies the necessity
of the static graph constraint component to the RE-GCN model. The static
information can be seen as the background knowledge of the TKG. The entity type
and location information in the static graph enriches the evolutional
representations of entities and helps obtain better initial evolutional
representations of entities. Note that, even without the static information,
--st w. GT still outperforms the state-of-art RE-NET w. GT and RGCRN w.
GT.

\subsubsection{\bf Impact of the Time Gate Recurrent Component} --tg w. GT in
Tables~\ref{table:ablation_ent} and ~\ref{table:ablation_evt} denotes a variant
of RE-GCN directly using the evolutional representations at the last timestamp
as the input of the evolution unit at the current timestamp without the time
gate. It can observe that the performance of --tg w. GT decreases rapidly
when the historical KG sequence gets longer, as compared to RE-GCN w. GT, which
sufficiently indicates the necessity of the time gate recurrent component.
Actually, the time gate recurrent component helps RE-GCN capture the sequence
patterns by a deep-stacked GCN, which usually faces the over-smoothing and
vanishing gradient problems when the number of the layers becomes large.

\subsection{Case Study}
In order to show the structural dependencies among concurrent facts and the
sequential patterns across temporally adjacent facts learned by RE-GCN, we
illustrate in Table~\ref{table:detail analysis} three cases that RE-GCN gives
the answers top 1 scores from the test set of ICEWS18. The first case shows the
sequential pattern that \emph{(A, host a visit, B, t-2)} can lead to \emph{(A,
diplomatic cooperation, B, t)}. The second case shows that the sequential
pattern \emph{(A, Conduct bombing, B, t-2)}, \emph{(B, Make statement, C, t-1)}
and structural dependencies of \emph{C} at timestamp $t-1$ joint lead to the
final result. The third case illustrates the sequential pattern \emph{(A,
Demonstrate, B, t-2), (B, Endorse, C, t-1)} helps the relation prediction
\emph{(A, ?, C, t)}. By modeling the KG sequence as a whole, RE-GCN does not
omit useful information in the history.
\vspace{-3mm}

\subsection{Detailed Analysis}
In order to get insight into the performance of RE-GCN on different kinds of
data, we conduct detailed analysis on the validation set of ICEWS18. For entity
prediction, we split the validation set according to the number of the one-hop
neighbors of a given entity (0 (a), 1 (b), 2-3 (c), 4-10 (d), >10 (e)) and
whether the answer entity has direct interactions with the given entity (i.e.,
seen and unseen) at the latest $m$ ($m$=6) timestamps. For relation prediction,
we split the validation set according to the number of relations between two
given entities (0 (a), 1 (b), 2-3 (c), 4-10 (d), >10 (e)) and whether the answer
relation occurred between the given entities at the latest $m$ timestamps.
(i.e., seen and unseen). Table~\ref{table:detail analysis} shows the results of
RE-GCN with Hits@3 on each subset. For entity prediction, it can be observed
that the performance decreases when the number of the neighbors gets large and
RE-GCN gets better results in the subset where the two entities have seen each
other in the history. Interestingly, RE-GCN can even conduct predictions where
the subject entities have no history. A possible reason is that the static graph
and the shared initial evolutional representations already provide some
background knowledge and information out of the historical KG sequence. For
relation prediction, it can be seen that the performance decreases when the
number of relations is large. Table~\ref{table:detail analysis} also
demonstrates the repetitive facts account for a certain proportion in the
dataset, which further proves the necessity of the time gate recurrent component
in RE-GCN.

\section{Conclusions}

This paper proposed RE-GCN for temporal reasoning, which learns evolutional
representations of entities and relations by capturing the structural
dependencies among concurrent facts and the informative sequential patterns
across temporally adjacent facts. Moreover, it incorporates the static
properties of entities such as entity types into the evolutional
representations. Thus, temporal reasoning is conducted with various score
functions based on the evolutional representations at the final timestamps.
Experimental results on six benchmarks demonstrate the significant merits and
superiority of RE-GCN on two temporal reasoning tasks. By modeling the KG
sequence as a whole, RE-GCN enables 17 to 82 times speedup in entity
prediction comparing to RE-NET, the state-of-the-art baseline.

\section{Acknowledge}
The work is supported by the National Key Research and Development Program of
China under grant 2016YFB1000902, the National Natural Science Foundation of
China under grants U1911401, 62002341, 61772501, U1836206, 91646120, and
61722211, the GFKJ Innovation Program, Beijing Academy of Artificial
Intelligence under grant BAAI2019ZD0306, and the Lenovo-CAS Joint Lab
Youth Scientist Project.

\bibliographystyle{ACM-Reference-Format}
\bibliography{ergcn}


\begin{thebibliography}{44}

%%% the Web address.

\ifx \showCODEN    \undefined \def \showCODEN     #1{\unskip}     \fi
\ifx \showDOI      \undefined \def \showDOI       #1{#1}\fi
\ifx \showISBNx    \undefined \def \showISBNx     #1{\unskip}     \fi
\ifx \showISBNxiii \undefined \def \showISBNxiii  #1{\unskip}     \fi
\ifx \showISSN     \undefined \def \showISSN      #1{\unskip}     \fi
\ifx \showLCCN     \undefined \def \showLCCN      #1{\unskip}     \fi
\ifx \shownote     \undefined \def \shownote      #1{#1}          \fi
\ifx \showarticletitle \undefined \def \showarticletitle #1{#1}   \fi
\ifx \showURL      \undefined \def \showURL       {\relax}        \fi
% invisible to TeX
\providecommand\bibfield[2]{#2}
\providecommand\bibinfo[2]{#2}
\providecommand\natexlab[1]{#1}
\providecommand\showeprint[2][]{arXiv:#2}

\bibitem[\protect\citeauthoryear{Bollen, Mao, and Zeng}{Bollen
  et~al\mbox{.}}{2011}]%
        {bollen2011twitter}
\bibfield{author}{\bibinfo{person}{Johan Bollen}, \bibinfo{person}{Huina Mao},
  {and} \bibinfo{person}{Xiaojun Zeng}.} \bibinfo{year}{2011}\natexlab{}.
\newblock \showarticletitle{Twitter mood predicts the stock market}.
\newblock \bibinfo{journal}{\emph{Journal of computational science}}
  \bibinfo{volume}{2}, \bibinfo{number}{1} (\bibinfo{year}{2011}),
  \bibinfo{pages}{1--8}.
\newblock


\bibitem[\protect\citeauthoryear{Bordes, Usunier, Garcia-Duran, Weston, and
  Yakhnenko}{Bordes et~al\mbox{.}}{2013}]%
        {bordes2013translating}
\bibfield{author}{\bibinfo{person}{Antoine Bordes}, \bibinfo{person}{Nicolas
  Usunier}, \bibinfo{person}{Alberto Garcia-Duran}, \bibinfo{person}{Jason
  Weston}, {and} \bibinfo{person}{Oksana Yakhnenko}.}
  \bibinfo{year}{2013}\natexlab{}.
\newblock \showarticletitle{Translating embeddings for modeling
  multi-relational data}. In \bibinfo{booktitle}{\emph{Advances in neural
  information processing systems}}. \bibinfo{pages}{2787--2795}.
\newblock


\bibitem[\protect\citeauthoryear{Boschee, Lautenschlager, O’Brien, Shellman,
  Starz, and Ward}{Boschee et~al\mbox{.}}{2015}]%
        {boschee2015icews}
\bibfield{author}{\bibinfo{person}{Elizabeth Boschee},
  \bibinfo{person}{Jennifer Lautenschlager}, \bibinfo{person}{Sean O’Brien},
  \bibinfo{person}{Steve Shellman}, \bibinfo{person}{James Starz}, {and}
  \bibinfo{person}{Michael Ward}.} \bibinfo{year}{2015}\natexlab{}.
\newblock \showarticletitle{ICEWS coded event data}.
\newblock \bibinfo{journal}{\emph{Harvard Dataverse}}  \bibinfo{volume}{12}
  (\bibinfo{year}{2015}).
\newblock


\bibitem[\protect\citeauthoryear{Dasgupta, Ray, and Talukdar}{Dasgupta
  et~al\mbox{.}}{2018}]%
        {dasgupta2018hyte}
\bibfield{author}{\bibinfo{person}{Shib~Sankar Dasgupta},
  \bibinfo{person}{Swayambhu~Nath Ray}, {and} \bibinfo{person}{Partha
  Talukdar}.} \bibinfo{year}{2018}\natexlab{}.
\newblock \showarticletitle{Hyte: Hyperplane-based temporally aware knowledge
  graph embedding}. In \bibinfo{booktitle}{\emph{Proceedings of the 2018
  Conference on Empirical Methods in Natural Language Processing}}.
  \bibinfo{pages}{2001--2011}.
\newblock


\bibitem[\protect\citeauthoryear{Deng, Rangwala, and Ning}{Deng
  et~al\mbox{.}}{2020}]%
        {deng2020dynamic}
\bibfield{author}{\bibinfo{person}{Songgaojun Deng}, \bibinfo{person}{Huzefa
  Rangwala}, {and} \bibinfo{person}{Yue Ning}.}
  \bibinfo{year}{2020}\natexlab{}.
\newblock \showarticletitle{Dynamic Knowledge Graph based Multi-Event
  Forecasting}. In \bibinfo{booktitle}{\emph{Proceedings of the 26th ACM SIGKDD
  International Conference on Knowledge Discovery \& Data Mining}}.
  \bibinfo{pages}{1585--1595}.
\newblock


\bibitem[\protect\citeauthoryear{Dettmers, Minervini, Stenetorp, and
  Riedel}{Dettmers et~al\mbox{.}}{2018}]%
        {dettmers2018convolutional}
\bibfield{author}{\bibinfo{person}{Tim Dettmers}, \bibinfo{person}{Pasquale
  Minervini}, \bibinfo{person}{Pontus Stenetorp}, {and}
  \bibinfo{person}{Sebastian Riedel}.} \bibinfo{year}{2018}\natexlab{}.
\newblock \showarticletitle{Convolutional 2d knowledge graph embeddings}. In
  \bibinfo{booktitle}{\emph{Thirty-Second AAAI Conference on Artificial
  Intelligence}}.
\newblock


\bibitem[\protect\citeauthoryear{Ding, Han, Ma, and Tresp}{Ding
  et~al\mbox{.}}{2021}]%
        {ding2021temporal}
\bibfield{author}{\bibinfo{person}{Zifeng Ding}, \bibinfo{person}{Zhen Han},
  \bibinfo{person}{Yunpu Ma}, {and} \bibinfo{person}{Volker Tresp}.}
  \bibinfo{year}{2021}\natexlab{}.
\newblock \showarticletitle{Temporal Knowledge Graph Forecasting with Neural
  ODE}.
\newblock \bibinfo{journal}{\emph{arXiv preprint arXiv:2101.05151}}
  (\bibinfo{year}{2021}).
\newblock


\bibitem[\protect\citeauthoryear{Esteban, Tresp, Yang, Baier, and
  Krompa{\ss}}{Esteban et~al\mbox{.}}{2016}]%
        {esteban2016predicting}
\bibfield{author}{\bibinfo{person}{Crist{\'o}bal Esteban},
  \bibinfo{person}{Volker Tresp}, \bibinfo{person}{Yinchong Yang},
  \bibinfo{person}{Stephan Baier}, {and} \bibinfo{person}{Denis Krompa{\ss}}.}
  \bibinfo{year}{2016}\natexlab{}.
\newblock \showarticletitle{Predicting the co-evolution of event and knowledge
  graphs}. In \bibinfo{booktitle}{\emph{2016 19th International Conference on
  Information Fusion (FUSION)}}. IEEE, \bibinfo{pages}{98--105}.
\newblock


\bibitem[\protect\citeauthoryear{Garc{\'\i}a-Dur{\'a}n, Duman{\v{c}}i{\'c}, and
  Niepert}{Garc{\'\i}a-Dur{\'a}n et~al\mbox{.}}{2018}]%
        {garcia2018learning}
\bibfield{author}{\bibinfo{person}{Alberto Garc{\'\i}a-Dur{\'a}n},
  \bibinfo{person}{Sebastijan Duman{\v{c}}i{\'c}}, {and}
  \bibinfo{person}{Mathias Niepert}.} \bibinfo{year}{2018}\natexlab{}.
\newblock \showarticletitle{Learning sequence encoders for temporal knowledge
  graph completion}.
\newblock \bibinfo{journal}{\emph{arXiv preprint arXiv:1809.03202}}
  (\bibinfo{year}{2018}).
\newblock


\bibitem[\protect\citeauthoryear{Goel, Kazemi, Brubaker, and Poupart}{Goel
  et~al\mbox{.}}{2020}]%
        {goel2020diachronic}
\bibfield{author}{\bibinfo{person}{Rishab Goel}, \bibinfo{person}{Seyed~Mehran
  Kazemi}, \bibinfo{person}{Marcus Brubaker}, {and} \bibinfo{person}{Pascal
  Poupart}.} \bibinfo{year}{2020}\natexlab{}.
\newblock \showarticletitle{Diachronic embedding for temporal knowledge graph
  completion}. In \bibinfo{booktitle}{\emph{Proceedings of the AAAI Conference
  on Artificial Intelligence}}, Vol.~\bibinfo{volume}{34}.
  \bibinfo{pages}{3988--3995}.
\newblock


\bibitem[\protect\citeauthoryear{Gottschalk and Demidova}{Gottschalk and
  Demidova}{2018}]%
        {gottschalk2018eventkg}
\bibfield{author}{\bibinfo{person}{Simon Gottschalk} {and}
  \bibinfo{person}{Elena Demidova}.} \bibinfo{year}{2018}\natexlab{}.
\newblock \showarticletitle{Eventkg: A multilingual event-centric temporal
  knowledge graph}. In \bibinfo{booktitle}{\emph{European Semantic Web
  Conference}}. Springer, \bibinfo{pages}{272--287}.
\newblock


\bibitem[\protect\citeauthoryear{Gottschalk and Demidova}{Gottschalk and
  Demidova}{2019}]%
        {gottschalk2019eventkg}
\bibfield{author}{\bibinfo{person}{Simon Gottschalk} {and}
  \bibinfo{person}{Elena Demidova}.} \bibinfo{year}{2019}\natexlab{}.
\newblock \showarticletitle{EventKG--the hub of event knowledge on the web--and
  biographical timeline generation}.
\newblock \bibinfo{journal}{\emph{Semantic Web}} \bibinfo{number}{Preprint}
  (\bibinfo{year}{2019}), \bibinfo{pages}{1--32}.
\newblock


\bibitem[\protect\citeauthoryear{Han, Chen, Ma, and Tresp}{Han
  et~al\mbox{.}}{2020a}]%
        {han2020dyernie}
\bibfield{author}{\bibinfo{person}{Zhen Han}, \bibinfo{person}{Peng Chen},
  \bibinfo{person}{Yunpu Ma}, {and} \bibinfo{person}{Volker Tresp}.}
  \bibinfo{year}{2020}\natexlab{a}.
\newblock \showarticletitle{Dyernie: Dynamic evolution of riemannian manifold
  embeddings for temporal knowledge graph completion}.
\newblock \bibinfo{journal}{\emph{arXiv preprint arXiv:2011.03984}}
  (\bibinfo{year}{2020}).
\newblock


\bibitem[\protect\citeauthoryear{Han, Ma, Wang, G{\"u}nnemann, and Tresp}{Han
  et~al\mbox{.}}{2020b}]%
        {han2020graph}
\bibfield{author}{\bibinfo{person}{Zhen Han}, \bibinfo{person}{Yunpu Ma},
  \bibinfo{person}{Yuyi Wang}, \bibinfo{person}{Stephan G{\"u}nnemann}, {and}
  \bibinfo{person}{Volker Tresp}.} \bibinfo{year}{2020}\natexlab{b}.
\newblock \showarticletitle{Graph Hawkes Neural Network for Forecasting on
  Temporal Knowledge Graphs}.
\newblock \bibinfo{journal}{\emph{8th Automated Knowledge Base Construction
  (AKBC)}} (\bibinfo{year}{2020}).
\newblock


\bibitem[\protect\citeauthoryear{Jain, Rathi, Chakrabarti, et~al\mbox{.}}{Jain
  et~al\mbox{.}}{2020}]%
        {jain2020temporal}
\bibfield{author}{\bibinfo{person}{Prachi Jain}, \bibinfo{person}{Sushant
  Rathi}, \bibinfo{person}{Soumen Chakrabarti}, {et~al\mbox{.}}}
  \bibinfo{year}{2020}\natexlab{}.
\newblock \showarticletitle{Temporal Knowledge Base Completion: New Algorithms
  and Evaluation Protocols}. In \bibinfo{booktitle}{\emph{Proceedings of the
  2020 Conference on Empirical Methods in Natural Language Processing
  (EMNLP)}}. \bibinfo{pages}{3733--3747}.
\newblock


\bibitem[\protect\citeauthoryear{Jin, Qu, Jin, and Ren}{Jin
  et~al\mbox{.}}{2020}]%
        {jin2020Renet}
\bibfield{author}{\bibinfo{person}{Woojeong Jin}, \bibinfo{person}{Meng Qu},
  \bibinfo{person}{Xisen Jin}, {and} \bibinfo{person}{Xiang Ren}.}
  \bibinfo{year}{2020}\natexlab{}.
\newblock \showarticletitle{Recurrent Event Network: Autoregressive Structure
  Inference over Temporal Knowledge Graphs}. In
  \bibinfo{booktitle}{\emph{EMNLP}}.
\newblock


\bibitem[\protect\citeauthoryear{Jin, Zhang, Szekely, and Ren}{Jin
  et~al\mbox{.}}{2019}]%
        {jin2019recurrent}
\bibfield{author}{\bibinfo{person}{Woojeong Jin}, \bibinfo{person}{Changlin
  Zhang}, \bibinfo{person}{Pedro Szekely}, {and} \bibinfo{person}{Xiang Ren}.}
  \bibinfo{year}{2019}\natexlab{}.
\newblock \showarticletitle{Recurrent Event Network for Reasoning over Temporal
  Knowledge Graphs}.
\newblock \bibinfo{journal}{\emph{arXiv preprint arXiv:1904.05530}}
  (\bibinfo{year}{2019}).
\newblock


\bibitem[\protect\citeauthoryear{Kingma and Ba}{Kingma and Ba}{2014}]%
        {kingma2014adam}
\bibfield{author}{\bibinfo{person}{Diederik~P Kingma} {and}
  \bibinfo{person}{Jimmy Ba}.} \bibinfo{year}{2014}\natexlab{}.
\newblock \showarticletitle{Adam: A method for stochastic optimization}.
\newblock \bibinfo{journal}{\emph{arXiv preprint arXiv:1412.6980}}
  (\bibinfo{year}{2014}).
\newblock


\bibitem[\protect\citeauthoryear{Kipf and Welling}{Kipf and Welling}{2016}]%
        {kipf2016semi}
\bibfield{author}{\bibinfo{person}{Thomas~N Kipf} {and} \bibinfo{person}{Max
  Welling}.} \bibinfo{year}{2016}\natexlab{}.
\newblock \showarticletitle{Semi-supervised classification with graph
  convolutional networks}.
\newblock \bibinfo{journal}{\emph{arXiv preprint arXiv:1609.02907}}
  (\bibinfo{year}{2016}).
\newblock


\bibitem[\protect\citeauthoryear{Korkmaz, Cadena, Kuhlman, Marathe, Vullikanti,
  and Ramakrishnan}{Korkmaz et~al\mbox{.}}{2015}]%
        {korkmaz2015combining}
\bibfield{author}{\bibinfo{person}{Gizem Korkmaz}, \bibinfo{person}{Jose
  Cadena}, \bibinfo{person}{Chris~J Kuhlman}, \bibinfo{person}{Achla Marathe},
  \bibinfo{person}{Anil Vullikanti}, {and} \bibinfo{person}{Naren
  Ramakrishnan}.} \bibinfo{year}{2015}\natexlab{}.
\newblock \showarticletitle{Combining heterogeneous data sources for civil
  unrest forecasting}. In \bibinfo{booktitle}{\emph{Proceedings of the 2015
  IEEE/ACM International Conference on Advances in Social Networks Analysis and
  Mining 2015}}. \bibinfo{pages}{258--265}.
\newblock


\bibitem[\protect\citeauthoryear{Leblay and Chekol}{Leblay and Chekol}{2018}]%
        {leblay2018deriving}
\bibfield{author}{\bibinfo{person}{Julien Leblay} {and}
  \bibinfo{person}{Melisachew~Wudage Chekol}.} \bibinfo{year}{2018}\natexlab{}.
\newblock \showarticletitle{Deriving validity time in knowledge graph}. In
  \bibinfo{booktitle}{\emph{Companion Proceedings of the The Web Conference
  2018}}. International World Wide Web Conferences Steering Committee,
  \bibinfo{pages}{1771--1776}.
\newblock


\bibitem[\protect\citeauthoryear{Leetaru and Schrodt}{Leetaru and
  Schrodt}{2013}]%
        {leetaru2013gdelt}
\bibfield{author}{\bibinfo{person}{Kalev Leetaru} {and}
  \bibinfo{person}{Philip~A Schrodt}.} \bibinfo{year}{2013}\natexlab{}.
\newblock \showarticletitle{Gdelt: Global data on events, location, and tone,
  1979--2012}. In \bibinfo{booktitle}{\emph{ISA annual convention}},
  Vol.~\bibinfo{volume}{2}. Citeseer, \bibinfo{pages}{1--49}.
\newblock


\bibitem[\protect\citeauthoryear{Li, Muller, Thabet, and Ghanem}{Li
  et~al\mbox{.}}{2019}]%
        {li2019deepgcns}
\bibfield{author}{\bibinfo{person}{Guohao Li}, \bibinfo{person}{Matthias
  Muller}, \bibinfo{person}{Ali Thabet}, {and} \bibinfo{person}{Bernard
  Ghanem}.} \bibinfo{year}{2019}\natexlab{}.
\newblock \showarticletitle{Deepgcns: Can gcns go as deep as cnns?}. In
  \bibinfo{booktitle}{\emph{Proceedings of the IEEE International Conference on
  Computer Vision}}. \bibinfo{pages}{9267--9276}.
\newblock


\bibitem[\protect\citeauthoryear{Mahdisoltani, Biega, and
  Suchanek}{Mahdisoltani et~al\mbox{.}}{2014}]%
        {mahdisoltani2014yago3}
\bibfield{author}{\bibinfo{person}{Farzaneh Mahdisoltani},
  \bibinfo{person}{Joanna Biega}, {and} \bibinfo{person}{Fabian Suchanek}.}
  \bibinfo{year}{2014}\natexlab{}.
\newblock \showarticletitle{Yago3: A knowledge base from multilingual
  wikipedias}. In \bibinfo{booktitle}{\emph{7th biennial conference on
  innovative data systems research}}. CIDR Conference.
\newblock


\bibitem[\protect\citeauthoryear{Muthiah, Huang, Arredondo, Mares, Getoor,
  Katz, and Ramakrishnan}{Muthiah et~al\mbox{.}}{2015}]%
        {muthiah2015planned}
\bibfield{author}{\bibinfo{person}{Sathappan Muthiah}, \bibinfo{person}{Bert
  Huang}, \bibinfo{person}{Jaime Arredondo}, \bibinfo{person}{David Mares},
  \bibinfo{person}{Lise Getoor}, \bibinfo{person}{Graham Katz}, {and}
  \bibinfo{person}{Naren Ramakrishnan}.} \bibinfo{year}{2015}\natexlab{}.
\newblock \showarticletitle{Planned protest modeling in news and social media}.
  In \bibinfo{booktitle}{\emph{Twenty-Seventh IAAI Conference}}. Citeseer.
\newblock


\bibitem[\protect\citeauthoryear{Phillips, Dowling, Shaffer, Hodas, and
  Volkova}{Phillips et~al\mbox{.}}{2017}]%
        {phillips2017using}
\bibfield{author}{\bibinfo{person}{Lawrence Phillips}, \bibinfo{person}{Chase
  Dowling}, \bibinfo{person}{Kyle Shaffer}, \bibinfo{person}{Nathan Hodas},
  {and} \bibinfo{person}{Svitlana Volkova}.} \bibinfo{year}{2017}\natexlab{}.
\newblock \showarticletitle{Using social media to predict the future: a
  systematic literature review}.
\newblock \bibinfo{journal}{\emph{arXiv preprint arXiv:1706.06134}}
  (\bibinfo{year}{2017}).
\newblock


\bibitem[\protect\citeauthoryear{Sadeghian, Rodriguez, Wang, and
  Colas}{Sadeghian et~al\mbox{.}}{2016}]%
        {sadeghian2016temporal}
\bibfield{author}{\bibinfo{person}{Ali Sadeghian}, \bibinfo{person}{Miguel
  Rodriguez}, \bibinfo{person}{Daisy~Zhe Wang}, {and} \bibinfo{person}{Anthony
  Colas}.} \bibinfo{year}{2016}\natexlab{}.
\newblock \showarticletitle{Temporal reasoning over event knowledge graphs}. In
  \bibinfo{booktitle}{\emph{Workshop on Knowledge Base Construction, Reasoning
  and Mining}}.
\newblock


\bibitem[\protect\citeauthoryear{Schlichtkrull, Kipf, Bloem, Van Den~Berg,
  Titov, and Welling}{Schlichtkrull et~al\mbox{.}}{2018}]%
        {schlichtkrull2018modeling}
\bibfield{author}{\bibinfo{person}{Michael Schlichtkrull},
  \bibinfo{person}{Thomas~N Kipf}, \bibinfo{person}{Peter Bloem},
  \bibinfo{person}{Rianne Van Den~Berg}, \bibinfo{person}{Ivan Titov}, {and}
  \bibinfo{person}{Max Welling}.} \bibinfo{year}{2018}\natexlab{}.
\newblock \showarticletitle{Modeling relational data with graph convolutional
  networks}. In \bibinfo{booktitle}{\emph{European Semantic Web Conference}}.
  Springer, \bibinfo{pages}{593--607}.
\newblock


\bibitem[\protect\citeauthoryear{Seo, Defferrard, Vandergheynst, and
  Bresson}{Seo et~al\mbox{.}}{2018}]%
        {seo2018structured}
\bibfield{author}{\bibinfo{person}{Youngjoo Seo}, \bibinfo{person}{Micha{\"e}l
  Defferrard}, \bibinfo{person}{Pierre Vandergheynst}, {and}
  \bibinfo{person}{Xavier Bresson}.} \bibinfo{year}{2018}\natexlab{}.
\newblock \showarticletitle{Structured sequence modeling with graph
  convolutional recurrent networks}. In \bibinfo{booktitle}{\emph{International
  Conference on Neural Information Processing}}. Springer,
  \bibinfo{pages}{362--373}.
\newblock


\bibitem[\protect\citeauthoryear{Shang, Tang, Huang, Bi, He, and Zhou}{Shang
  et~al\mbox{.}}{2019}]%
        {shang2019end}
\bibfield{author}{\bibinfo{person}{Chao Shang}, \bibinfo{person}{Yun Tang},
  \bibinfo{person}{Jing Huang}, \bibinfo{person}{Jinbo Bi},
  \bibinfo{person}{Xiaodong He}, {and} \bibinfo{person}{Bowen Zhou}.}
  \bibinfo{year}{2019}\natexlab{}.
\newblock \showarticletitle{End-to-end structure-aware convolutional networks
  for knowledge base completion}. In \bibinfo{booktitle}{\emph{Proceedings of
  the AAAI Conference on Artificial Intelligence}}, Vol.~\bibinfo{volume}{33}.
  \bibinfo{pages}{3060--3067}.
\newblock


\bibitem[\protect\citeauthoryear{Signorini, Segre, and Polgreen}{Signorini
  et~al\mbox{.}}{2011}]%
        {signorini2011use}
\bibfield{author}{\bibinfo{person}{Alessio Signorini},
  \bibinfo{person}{Alberto~Maria Segre}, {and} \bibinfo{person}{Philip~M
  Polgreen}.} \bibinfo{year}{2011}\natexlab{}.
\newblock \showarticletitle{The use of Twitter to track levels of disease
  activity and public concern in the US during the influenza A H1N1 pandemic}.
\newblock \bibinfo{journal}{\emph{PloS one}} \bibinfo{volume}{6},
  \bibinfo{number}{5} (\bibinfo{year}{2011}), \bibinfo{pages}{e19467}.
\newblock


\bibitem[\protect\citeauthoryear{Sun, Deng, Nie, and Tang}{Sun
  et~al\mbox{.}}{2018}]%
        {sun2018rotate}
\bibfield{author}{\bibinfo{person}{Zhiqing Sun}, \bibinfo{person}{Zhi-Hong
  Deng}, \bibinfo{person}{Jian-Yun Nie}, {and} \bibinfo{person}{Jian Tang}.}
  \bibinfo{year}{2018}\natexlab{}.
\newblock \showarticletitle{RotatE: Knowledge Graph Embedding by Relational
  Rotation in Complex Space}. In \bibinfo{booktitle}{\emph{International
  Conference on Learning Representations}}.
\newblock


\bibitem[\protect\citeauthoryear{Trivedi, Dai, Wang, and Song}{Trivedi
  et~al\mbox{.}}{2017}]%
        {trivedi2017know}
\bibfield{author}{\bibinfo{person}{Rakshit Trivedi}, \bibinfo{person}{Hanjun
  Dai}, \bibinfo{person}{Yichen Wang}, {and} \bibinfo{person}{Le Song}.}
  \bibinfo{year}{2017}\natexlab{}.
\newblock \showarticletitle{Know-evolve: Deep temporal reasoning for dynamic
  knowledge graphs}. In \bibinfo{booktitle}{\emph{Proceedings of the 34th
  International Conference on Learning-Volume 70}}. JMLR. org,
  \bibinfo{pages}{3462--3471}.
\newblock


\bibitem[\protect\citeauthoryear{Trivedi, Farajtabar, Biswal, and Zha}{Trivedi
  et~al\mbox{.}}{2018}]%
        {trivedi2018dyrep}
\bibfield{author}{\bibinfo{person}{Rakshit Trivedi}, \bibinfo{person}{Mehrdad
  Farajtabar}, \bibinfo{person}{Prasenjeet Biswal}, {and}
  \bibinfo{person}{Hongyuan Zha}.} \bibinfo{year}{2018}\natexlab{}.
\newblock \showarticletitle{Dyrep: Learning representations over dynamic
  graphs}.
\newblock  (\bibinfo{year}{2018}).
\newblock


\bibitem[\protect\citeauthoryear{Trouillon, Welbl, Riedel, Gaussier, and
  Bouchard}{Trouillon et~al\mbox{.}}{2016}]%
        {trouillon2016complex}
\bibfield{author}{\bibinfo{person}{Th{\'e}o Trouillon},
  \bibinfo{person}{Johannes Welbl}, \bibinfo{person}{Sebastian Riedel},
  \bibinfo{person}{{\'E}ric Gaussier}, {and} \bibinfo{person}{Guillaume
  Bouchard}.} \bibinfo{year}{2016}\natexlab{}.
\newblock \showarticletitle{Complex embeddings for simple link prediction}. In
  \bibinfo{booktitle}{\emph{International Conference on Machine Learning}}.
  \bibinfo{pages}{2071--2080}.
\newblock


\bibitem[\protect\citeauthoryear{Vashishth, Sanyal, Nitin, and
  Talukdar}{Vashishth et~al\mbox{.}}{2019}]%
        {vashishth2019composition}
\bibfield{author}{\bibinfo{person}{Shikhar Vashishth}, \bibinfo{person}{Soumya
  Sanyal}, \bibinfo{person}{Vikram Nitin}, {and} \bibinfo{person}{Partha
  Talukdar}.} \bibinfo{year}{2019}\natexlab{}.
\newblock \showarticletitle{Composition-based Multi-Relational Graph
  Convolutional Networks}. In \bibinfo{booktitle}{\emph{International
  Conference on Learning Representations}}.
\newblock


\bibitem[\protect\citeauthoryear{Wu, Cao, Cheung, and Hamilton}{Wu
  et~al\mbox{.}}{2020}]%
        {wu2020temp}
\bibfield{author}{\bibinfo{person}{Jiapeng Wu}, \bibinfo{person}{Meng Cao},
  \bibinfo{person}{Jackie Chi~Kit Cheung}, {and} \bibinfo{person}{William~L
  Hamilton}.} \bibinfo{year}{2020}\natexlab{}.
\newblock \showarticletitle{TeMP: Temporal Message Passing for Temporal
  Knowledge Graph Completion}.
\newblock \bibinfo{journal}{\emph{arXiv preprint arXiv:2010.03526}}
  (\bibinfo{year}{2020}).
\newblock


\bibitem[\protect\citeauthoryear{Wu, Khan, Gao, and Li}{Wu
  et~al\mbox{.}}{2019}]%
        {wu2019efficiently}
\bibfield{author}{\bibinfo{person}{Tianxing Wu}, \bibinfo{person}{Arijit Khan},
  \bibinfo{person}{Huan Gao}, {and} \bibinfo{person}{Cheng Li}.}
  \bibinfo{year}{2019}\natexlab{}.
\newblock \showarticletitle{Efficiently embedding dynamic knowledge graphs}.
\newblock \bibinfo{journal}{\emph{arXiv preprint arXiv:1910.06708}}
  (\bibinfo{year}{2019}).
\newblock


\bibitem[\protect\citeauthoryear{Xu, Wang, Chen, and Li}{Xu
  et~al\mbox{.}}{2015}]%
        {xu2015empirical}
\bibfield{author}{\bibinfo{person}{Bing Xu}, \bibinfo{person}{Naiyan Wang},
  \bibinfo{person}{Tianqi Chen}, {and} \bibinfo{person}{Mu Li}.}
  \bibinfo{year}{2015}\natexlab{}.
\newblock \showarticletitle{Empirical evaluation of rectified activations in
  convolutional network}.
\newblock \bibinfo{journal}{\emph{arXiv preprint arXiv:1505.00853}}
  (\bibinfo{year}{2015}).
\newblock


\bibitem[\protect\citeauthoryear{Xu, Nayyeri, Alkhoury, Yazdi, and Lehmann}{Xu
  et~al\mbox{.}}{2020}]%
        {xu2020temporal}
\bibfield{author}{\bibinfo{person}{Chenjin Xu}, \bibinfo{person}{Mojtaba
  Nayyeri}, \bibinfo{person}{Fouad Alkhoury}, \bibinfo{person}{Hamed Yazdi},
  {and} \bibinfo{person}{Jens Lehmann}.} \bibinfo{year}{2020}\natexlab{}.
\newblock \showarticletitle{Temporal Knowledge Graph Completion Based on Time
  Series Gaussian Embedding}. In \bibinfo{booktitle}{\emph{International
  Semantic Web Conference}}. Springer, \bibinfo{pages}{654--671}.
\newblock


\bibitem[\protect\citeauthoryear{Yang, Yih, He, Gao, and Deng}{Yang
  et~al\mbox{.}}{2014}]%
        {yang2014embedding}
\bibfield{author}{\bibinfo{person}{Bishan Yang}, \bibinfo{person}{Wen-tau Yih},
  \bibinfo{person}{Xiaodong He}, \bibinfo{person}{Jianfeng Gao}, {and}
  \bibinfo{person}{Li Deng}.} \bibinfo{year}{2014}\natexlab{}.
\newblock \showarticletitle{Embedding entities and relations for learning and
  inference in knowledge bases}.
\newblock \bibinfo{journal}{\emph{arXiv preprint arXiv:1412.6575}}
  (\bibinfo{year}{2014}).
\newblock


\bibitem[\protect\citeauthoryear{Ye, Li, Fang, Zang, and Wang}{Ye
  et~al\mbox{.}}{2019}]%
        {ye2019vectorized}
\bibfield{author}{\bibinfo{person}{Rui Ye}, \bibinfo{person}{Xin Li},
  \bibinfo{person}{Yujie Fang}, \bibinfo{person}{Hongyu Zang}, {and}
  \bibinfo{person}{Mingzhong Wang}.} \bibinfo{year}{2019}\natexlab{}.
\newblock \showarticletitle{A vectorized relational graph convolutional network
  for multi-relational network alignment}. In
  \bibinfo{booktitle}{\emph{Proceedings of the Twenty-Eighth International
  Joint Conference on Artificial Intelligence, IJCAI-19}}.
  \bibinfo{pages}{4135--4141}.
\newblock


\bibitem[\protect\citeauthoryear{Zhu, Chen, Fan, Cheng, and Zhan}{Zhu
  et~al\mbox{.}}{2020}]%
        {zhu2020learning}
\bibfield{author}{\bibinfo{person}{Cunchao Zhu}, \bibinfo{person}{Muhao Chen},
  \bibinfo{person}{Changjun Fan}, \bibinfo{person}{Guangquan Cheng}, {and}
  \bibinfo{person}{Yan Zhan}.} \bibinfo{year}{2020}\natexlab{}.
\newblock \showarticletitle{Learning from History: Modeling Temporal Knowledge
  Graphs with Sequential Copy-Generation Networks}.
\newblock \bibinfo{journal}{\emph{arXiv preprint arXiv:2012.08492}}
  (\bibinfo{year}{2020}).
\newblock


\bibitem[\protect\citeauthoryear{Zou}{Zou}{2020}]%
        {zou2020survey}
\bibfield{author}{\bibinfo{person}{Xiaohan Zou}.}
  \bibinfo{year}{2020}\natexlab{}.
\newblock \showarticletitle{A survey on application of knowledge graph}. In
  \bibinfo{booktitle}{\emph{Journal of Physics: Conference Series}},
  Vol.~\bibinfo{volume}{1487}. IOP Publishing, \bibinfo{pages}{012016}.
\newblock


\end{thebibliography}


\end{document}